\pdfoutput=1

\documentclass[11pt]{article}

\usepackage[]{acl}

\usepackage{times}
\usepackage{latexsym}

\usepackage[T1]{fontenc}

\usepackage[utf8]{inputenc}

\usepackage{microtype}
\usepackage{algorithm}
\usepackage{algcompatible,amsmath}
\usepackage{graphicx}
\usepackage{url}
\usepackage{multirow}
\usepackage{comment}
\usepackage{amssymb}
\usepackage{pifont}
\usepackage{balance}
\usepackage{mathtools}
\usepackage{xcolor}
\usepackage{colortbl}
\usepackage{bbm}
\usepackage{enumitem}
\usepackage{marvosym}
\usepackage{textcomp}
\usepackage{array}
\usepackage{booktabs}
\usepackage{tabularx}
\usepackage{CJKutf8}
\usepackage{algorithmicx}  
\usepackage{algpseudocode} 
\usepackage{makecell}
\usepackage{siunitx}

\usepackage{float}

\usepackage{newfloat}
\usepackage{listings}
\usepackage{mdframed} 
\usepackage{tikz}

\usepackage[most]{tcolorbox}
\usepackage{listings} 

\mdfsetup{
  skipabove=0pt,
  skipbelow=4pt,
  innertopmargin=5pt, 
  innerbottommargin=5pt, 
  linewidth=0pt,  
  backgroundcolor=lightgray,
}
\newcolumntype{C}{>{\centering\arraybackslash}p{1em}}

\floatstyle{ruled}
\newfloat{listing}{tb}{lst}{}
\floatname{listing}{Listing}

\newcommand{\squishlist}{
\begin{list}{$\bullet$}
{ \usecounter{Lcount}
\setlength{\itemsep}{0pt}
\setlength{\parsep}{0pt}
\setlength{\topsep}{0pt}
\setlength{\partopsep}{0pt}
\setlength{\leftmargin}{2em}
\setlength{\labelwidth}{1.5em}
\setlength{\labelsep}{0.5em} } }
\newcommand{\squishend}{
\end{list} }
\makeatletter
\newcommand{\thickhline}{%
    \noalign {\ifnum 0=`}\fi \hrule height 2pt
    \futurelet \reserved@a \@xhline
}
\newcommand{\modelname}{\textsc{LLMArena}}
\newcommand{\trueskill}{TrueSkill\texttrademark~}

\title{\modelname: Assessing Capabilities of Large Language Models in Dynamic Multi-Agent Environments}

\author{Junzhe Chen$^{1*}$, Xuming Hu$^{2\dagger}$, Shuodi Liu$^{3}$, Shiyu Huang$^{4}$,\\ \textbf{Wei-Wei Tu}$^{4}$, \textbf{Zhaofeng He}$^{3}$, \textbf{Lijie Wen}$^{1\dagger}$\\
  $^1$Tsinghua University, $^2$The Hong Kong University of Science and Technology (Guangzhou),\\
  $^3$Beijing University of Posts and Telecommunications, $^4$4Paradigm Inc.\\ \texttt{chenjz20@mails.tsinghua.edu.cn,}\\ \texttt{xuminghu97@gmail.com, wenlj@tsinghua.edu.cn}
  }

\begin{document}
\maketitle
{
\let\thefootnote\relax\footnotetext{
$^*$ Work was done during the intern at 4Paradigm Inc..}
\let\thefootnote\relax\footnotetext{
$^\dagger$ Corresponding authors. }
}
\begin{abstract}
Recent advancements in large language models (LLMs) have revealed their potential for achieving autonomous agents possessing human-level intelligence. However, existing benchmarks for evaluating LLM Agents either use static datasets, potentially leading to data leakage, or focus only on single-agent scenarios, overlooking the complexities of multi-agent interactions. There is a lack of a benchmark that evaluates the diverse capabilities of LLM agents in multi-agent, dynamic environments. To this end, we introduce \modelname, a novel and easily extensible framework for evaluating the diverse capabilities of LLM in multi-agent dynamic environments. \modelname~ encompasses seven distinct gaming environments, employing \trueskill scoring to assess crucial abilities in LLM agents, including spatial reasoning, strategic planning, numerical reasoning, risk assessment, communication, opponent modeling, and team collaboration. We conduct an extensive experiment and human evaluation among different sizes and types of LLMs, showing that LLMs still have a significant journey ahead in their development towards becoming fully autonomous agents, especially in opponent modeling and team collaboration. We hope \modelname~ could guide future research towards enhancing these capabilities in LLMs, ultimately leading to more sophisticated and practical applications in dynamic, multi-agent settings. The code and data will be available.
\end{abstract}
\section{Introduction}
Recent large language models (LLMs) have greatly promoted the progress of the field of natural language processing (NLP) due to their excellent zero-shot capabilities in various downstream tasks~\cite{kojima2022large,yang2023harnessing,llama,llama2,chatgpt,gpt4}. LLMs not only excel in comprehending and generating intricate text but also demonstrate remarkable adaptability, even without training tailored to specific tasks~\cite{wei2023zero,guo2023images}. They possess the ability to analyze, strategize, and respond effectively to unfamiliar scenarios using minimal guiding prompts. This has opened new avenues for researchers to consider LLMs as autonomous agents, providing automated assistance in complex real-world tasks such as software development and knowledge integration\cite{qian2023communicative,pan2024unifying}.

To better understand the skills needed by LLM as an agent, researchers are now concentrating on creating and using various scenarios to assess LLM's performance in specific settings~\cite{liu2023agentbench,chen2023agentverse,wu2023smartplay,li2023camel,yao2022webshop}. For instance, AgentBench introduced eight distinct environments, such as operating systems and databases, to examine LLM's proficiency in code generation and lateral thinking. Similarly, SmartPlay tests LLM in a gaming context as a solo agent to gauge its reasoning and strategic planning abilities. Nonetheless, these benchmarks have their constraints. Firstly, static datasets used in benchmarks like AgentBench might lead to issues such as data leakage and overfitting, as LLMs could have already encountered this data during their pre-training phase~\cite{zhou2023don}. Secondly, evaluation systems that focus solely on a single agent also have evident limitations, particularly in analyzing an agent's performance in intricate and dynamically interactive settings. Such single-agent environments tend to concentrate on the interactions between the agent and its environment, thereby overlooking the complex dynamics of interactions that occur among multiple agents in a shared environment~\cite{foster2023complexity}.
\begin{figure*}
  \centering
  \includegraphics[width=\textwidth]{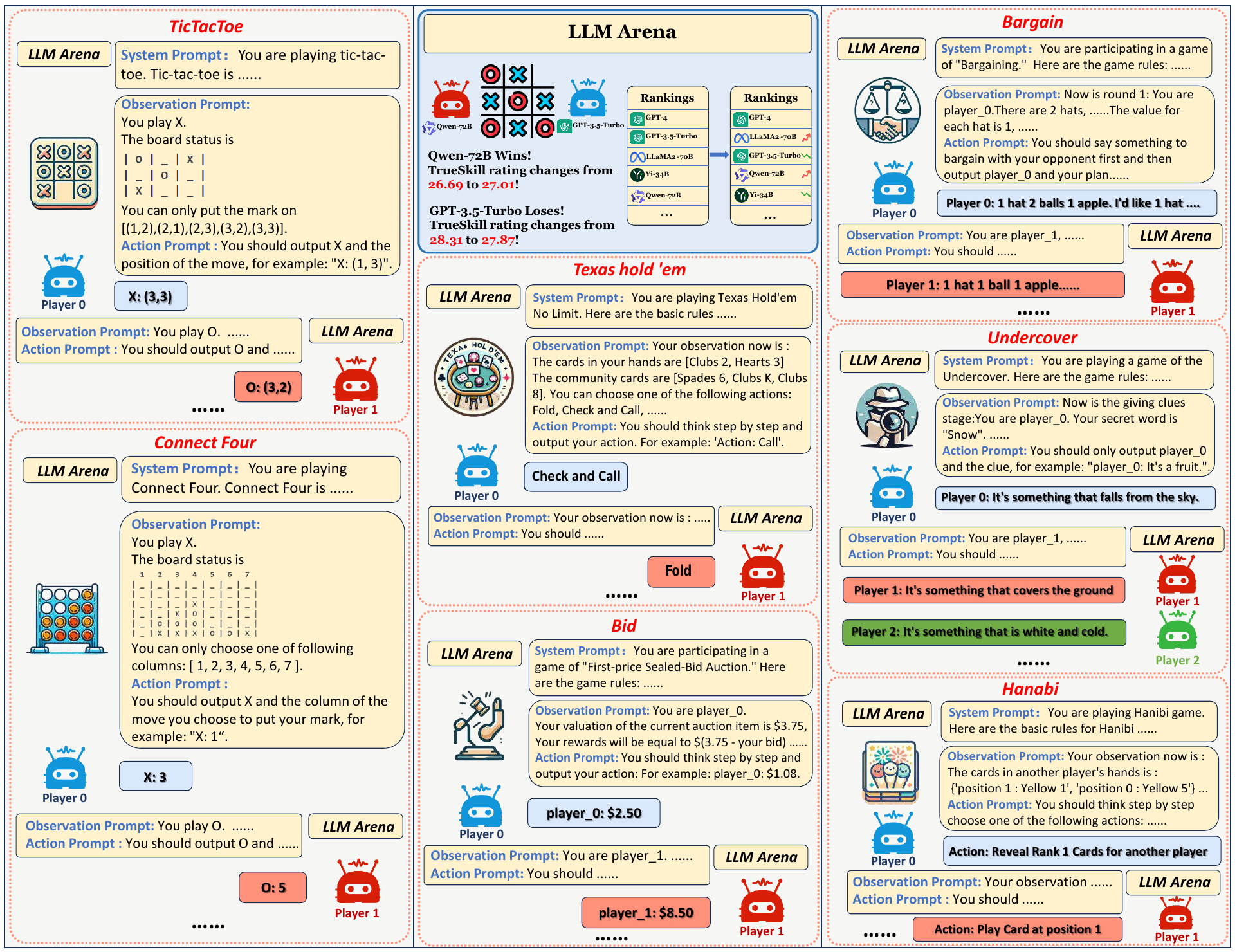}
  \caption{Examples of seven different game environments in \modelname.}
  \label{figure:arena}
  \vspace{-5mm}
\end{figure*}
To this end, we proposed \modelname, a dynamic evaluation benchmark specially designed for multi-agent interaction. As illustrated in Figure~\ref{figure:arena}, \modelname~ covers seven different types of dynamic, multi-agent game environments. For example, the Texas Hold'em poker game environment automatically generates a new hand for each game, and the complexity of each round's state increases exponentially. This dynamic nature of the game serves to mitigate the issue of data leakage. In the Undercover environment, the LLM agents need to point out the ``undercover'' during the communication process with other agents, which measures the LLM agents' communication ability and opponent modeling ability when facing different agents. Based on these environments, we can comprehensively evaluate LLM's spatial comprehension, strategic planning, numerical reasoning, risk assessment, communication, opponent modeling, and team collaboration capabilities in the dynamic multi-agent environment.

To more accurately evaluate LLM's performance in these multi-agent environments, we adopted the \trueskill~\cite{herbrich2006trueskill} scoring system, which can better help \modelname~ evaluate other metrics besides win rate, such as the relative skill level between agents, to provide a more comprehensive assessment under different scenarios and opponent configurations.

We conduct an extensive experiment and human evaluation among 14 different sizes and types of LLMs. The results indicated significant potential for improvement in LLM's teamwork and opponent modeling skills within multi-agent environments.  Based on these results, we hope \modelname~ can inspire future research focused on enhancing the core capabilities of LLM in multi-agent settings to promote the widespread use of LLM agents in real-world applications.

\section{Benchmark Detail}
In this section, we provide a detailed overview of LLMArena's composition, which includes its seven distinct game environments, evaluation metrics, and the methodologies used for assessment.
\subsection{Benchmark Overview}
\label{overview}
Figure \ref{figure:arena} illustrates the seven distinct environments within \modelname. Within each specific environment, LLMs may need to utilize a unique combination of capabilities to effectively tackle these challenges. For instance, in the Undercover environment, LLMs are required to possess a multifaceted skill set, including opponent modeling, communicating effectively, and team collaboration. The absence of any one of these abilities could result in ultimate failure in this environment.
\subsection{Benchmark Construction}
To ensure user-friendliness and robust scalability in \modelname, we have developed the environment using PettingZoo~\cite{terry2021pettingzoo} as a foundation. This approach enables subsequent researchers to effortlessly integrate new environments into this framework. By simply creating environments that adhere to the PettingZoo interface specifications\footnote{\url{https://pettingzoo.farama.org/content/environment_creation/}}, they can seamlessly add these to \modelname~for evaluating the diverse capabilities of LLMs. Every environment within \modelname~offers system prompts that outline the game rules and prompt templates designed to guide LLM agents in gameplay. These templates include information on the current status of the game, historical data, and a range of optional actions, facilitating a more structured and informed gaming experience for the LLM agents. All the prompts are shown in Appendix \ref{appendix:prompt}.
\subsection{Metrics}
\label{metrics}
In this section, we outline the evaluation metrics used by \modelname, followed by a comparative analysis with metrics utilized in prior research.
\paragraph{\trueskill} \trueskill is a scoring system developed by~\citet{herbrich2006trueskill} that evaluates multiple agents' skill levels in competitive gaming. In contrast to static, opponent-independent metrics employed in previous research, such as Win Rate and Completion Rate~\cite{wu2023smartplay,xu2023magic}, \trueskill offers a more nuanced assessment. It goes beyond merely accounting for wins and losses by also considering the quality of the game and the skill disparities between players. Within the \trueskill framework, defeating higher-skilled opponents yields more points compared to victories over lower-skilled adversaries. This approach stands in stark contrast to simple win rate calculations, which do not account for the skill level of the opponent. Such opponent-related metrics provide an accurate evaluation of an agent's real competencies in a multi-agent environment.
\paragraph{Reward} Currently, a substantial amount of research employs expert LLMs to assess the capabilities of other LLMs, yet this approach frequently lacks interpretability~ \cite{wang2023large}. \modelname~ addresses the gap by leveraging the concept of reward from reinforcement learning to measure the quality of each action performed by agents. The method allows for a quantitative and interpretable evaluation of LLMs' capabilities, offering a more nuanced and understandable analysis.
\subsection{Environment Settings}
\label{settings}
In this section, we give a comprehensive explanation of the diverse game environments included in the \modelname~ benchmark, highlighting the specific challenges that language models encounter while engaging with these environments.
\paragraph{TicTacToe:} TicTacToe\footnote{\url{https://pettingzoo.farama.org/environments/classic/tictactoe/}} is a strategy game where two LLM agents alternate in placing their marks on a 3$\times$3 grid. Victory is achieved when one agent successfully aligns three of their marks in a row, either horizontally, vertically, or diagonally. The game results in a draw if all cells are filled without any player achieving a winning alignment. This game examines LLM's \textit{strategic planning} and \textit{spatial reasoning} capabilities. We adopt the \trueskill ratings as the evaluation metrics.
\paragraph{ConnectFour}
As a more complex case of the board game, ConnectFour\footnote{\url{https://pettingzoo.farama.org/environments/classic/connect_four/}} uses a 6$\times$7 chessboard. Here, two LLM agents alternately choose one of the seven columns with available empty cells. Each piece placed will descend to the lowest available space in the selected column, either resting at the bottom or atop a previously placed token. An agent secures victory by aligning four of their pieces either horizontally, vertically, or diagonally. The game concludes in a draw if all cells are filled without a winner. Similar to TicTacToe, in ConnectFour, the LLMs primarily encounter challenges in \textit{strategic reasoning} and \textit{spatial reasoning}. We adopt the \trueskill ratings as the evaluation metrics.
\paragraph{Texas Hold’em}
Texas Hold'em\footnote{\url{https://pettingzoo.farama.org/environments/classic/texas_holdem_no_limit/}} is a well-known card game. In \modelname, each game is played by two LLM agents. At the beginning of each game, players are dealt two private cards, known as hole cards. Subsequently, five community cards are dealt across three stages - the flop, the turn, and the river. Players aim to construct the best possible five-card hand using a combination of their hole cards and the community cards. Throughout the game, there are four betting rounds. In each round, players have the option to choose from the following actions: Fold, Check \& Call, Raise Half Pot, Raise Full Pot, and All-In. This game environment demands a range of capabilities from LLMs, including \textit{numerical reasoning}, \textit{opponent modeling}, and \textit{risk assessment}. We adopt the \trueskill ratings as the evaluation metrics.
\paragraph{Undercover}
Undercover is a party game, where players are divided into two groups: undercover and non-undercover. Two groups of players each receive a pair of similar but distinct words. The game has two stages per round: a communication stage where players describe their words, followed by a voting stage to identify an undercover. At this time, if all undercovers are eliminated, the non-undercovers win, otherwise the next round will proceed. In \modelname, 5 LLM agents play in each game, which includes four non-undercover agents using GPT-3.5-Turbo and one undercover LLM undergoing evaluation. The game lasts up to two rounds, and if the undercovers remain undetected, they win. Given the superior capabilities of GPT-4, employing it as the non-undercover agent would render it nearly impossible for other models to win. Therefore, we choose GPT-3.5-Turbo as non-undercover agents. This game tests an LLM's proficiency in \textit{communication}, \textit{opponent modeling}, and \textit{team collaboration}. The evaluation metric used here is the winning rate of each LLM when playing as the undercover against GPT-3.5-Turbo.
\paragraph{Bargain}
Bargain~\cite{lewis2017deal} is a semi-cooperative game where two LLM agents are required to devise a distribution strategy for a pool of items. Each item holds a distinct, undisclosed value for both agents. The game aims to reach a consensus on the allocation. When an agreement on the allocation is achieved, the agent with the maximum total value from the items secured emerges victorious. Bargain requires LLM to have the ability of \textit{numerical reasoning}, \textit{communication}, and \textit{opponent modeling}. We adopt the \trueskill ratings as the evaluation metrics.
\paragraph{Bid}
A first-price sealed-bid auction\footnote{\url{https://en.wikipedia.org/wiki/First-price_sealed-bid_auction}} is a widely used auction format in which all participants place bids without knowledge of each other's bids, and the highest bidder wins. In \modelname, two LLM agents are participating in each auction. Each agent receives a random inner valuation of the item. The winning LLM in the auction will receive their inner valuation minus the actual bid as a reward. This environment evaluates the LLM's \textit{numerical reasoning} and \textit{opponent modeling} abilities. We measure the LLM's performance using the average reward it achieves as the evaluation metric.
\paragraph{Hanabi} Hanabi is a cooperative card game wherein two LLM agents can only view each other's hand of cards. They use info tokens to ``reveal cards'' to each other, ``discard cards'' to acquire more information tokens, and ``play cards'' in a certain order to form fireworks. This requires \textit{team collaboration}, \textit{strategic planning}, and \textit{numerical reasoning} skills from the LLMs. Upon successfully setting up a firework, both LLM agents are rewarded with points equivalent to the sum of the card values in the firework. In \modelname, we form the same two LLMs into a cooperation team, and we employ the average reward obtained by various LLM teams as an evaluation metric.

\section{Experiments}
\begin{table*}[t]
\centering
\scalebox{0.79}{
\begin{tabular}{ccccccccc}
\toprule
LLMs                  & TicTacToe & ConnectFour & Texas Hold'em & Bid    & Bargain & Undercover & Hanabi & Average   \\
\midrule
\midrule
GPT-4     & 100.00    & 100.00      & 100.00           & 100.00& 100.00  & 100.00     & 100.00 & 100.00\\
GPT-3.5-Turbo     & 82.81     & 96.30       & 81.23            & 84.17& 99.26   &      92.59      &   84.13     &  88.64\\
\midrule
Qwen-72B-Chat         & 90.08     & 80.30       & 94.88            & 84.42& 92.95   &      62.96      &   89.01     &   84.94\\
Llama-2-70B   & 82.21     & 95.90       & 98.42            & 94.22& 88.63   &      85.19      &   80.35     &   89.27\\
Agentlm-70B           & 81.44     & 83.62       & 77.35            & 86.10& 98.33   &      92.59      &   79.58     &   85.57\\
DeepSeek-67B-Chat & 68.00     & 88.88       & 96.64            & 20.30& 89.02   &      74.07      &   68.69     &    72.23\\
\midrule
SUS-Chat-34B          & 79.75     & 96.30       & 84.38            & 66.17& 74.53   &     88.89       &   86.33     &   82.34\\
Yi-34B-Chat           & 78.23     & 95.36       & 86.06            & 88.77& 74.18   &     96.30    &   34.30     &     79.03\\
\midrule
Qwen-14B-Chat         & 86.40     & 72.67       & 90.56            & 86.82& 91.13   &      88.89      &  71.70      &   84.02\\
WizardLM-13B      & 79.14     & 83.35       & 74.26            & 35.28& 91.28   &     48.15       &  54.23      &   66.53\\
AgentLM-13B           & 78.89     & 89.23       & 86.85            & 30.39& 94.93   &     62.96       &   22.09     &   66.48\\
DeepSeek-7B-Chat  & 81.17     & 73.06       & 96.10            & 14.53& 92.56   &     74.07&  0.00      &    61.64\\
AgentLM-7B            & 68.29     & 90.35       & 85.14            & 0.04& 83.53&     66.67       &  98.20      &   70.32\\
Vicuna-7B             & 64.95     & 80.12       & 90.46            & 68.94& 86.33   &      62.96      &  87.37      &    77.30\\
\midrule
Average  &80.10     & 87.53       & 88.74           & 61.44  & 89.76   &      78.31      &  68.28      &    79.16   \\
\bottomrule
\end{tabular}
}
\caption{Normalized scores of 14 different LLMs in 7 environments.}
\vspace{-5mm}
\label{main-result}
\end{table*}


\subsection{Experiments Setup}
We conduct experiments with the closed-source GPT family LLM by directly calling its API. As for the open-source LLM, we deploy it locally and encapsulate it as an OpenAI API call. To ensure reproducibility, we set the temperature of all LLMs to 0. In all environments except Undercover, we conducted \modelname~runs until the \trueskill ratings between models reached convergence, with over 50 games played per environment per model. For the Undercover environment, we played 100 games of each LLM as an undercover agent in a game with four GPT-3.5-turbo as non-undercover agents. Initially, all LLM agents started with a \trueskill~score of 25.0 and a variance of 8.33.

\subsection{Main Result}
In Table \ref{main-result}, we present the relative normalized scores for 14 different LLMs across 7 environments within \modelname. The original data before normalization can be found in Appendix~\ref{origin}. This table allows us to draw the following conclusions.

$\bullet$ As the scale of the model parameters increases, there is a noticeable improvement in the capabilities of LLMs. For instance, LLMs with around 70B parameters exhibit an average performance of 82.87. LLMs with approximately 30B and 10B parameters average 80.68 and 71.05 in performance, respectively. It's noteworthy that the performance enhancement observed when escalating the model size from 10B to 30B parameters (+9.63) is more pronounced than the improvement seen from 30B to 70B parameters (+2.19).

$\bullet$ While it's generally observed that LLMs with a larger scale of parameters tend to outperform their counterparts with fewer parameters within the same series, exceptions can arise in specific environments. A case in point is the performance of Qwen-72B in the Undercover environment is surprisingly 25.93 points lower than that of Qwen-14B, which notably diverges from this trend. 

$\bullet$ The performance of LLMs in both the Bid and Hanabi environments significantly lags behind the average of all 7 environments, with deficits of 17.72 and 10.88 points, respectively. Smaller-scale LLMs ($\sim$10B) fared even worse, scoring 31.71 and 15.45 points below the average in these environments. This underscores that tasks involving numerical reasoning, opponent modeling, and team collaboration pose significant challenges for LLMs.

$\bullet$ There remains a significant disparity between GPT-4 and other models. GPT-4 has attained SOTA performance across all evaluated tasks, outperforming other models by an average margin of 16.76.

\section{Analysis}
\label{analysis}
\begin{table*}[t]

\centering
\scalebox{0.71}{
\begin{tabular}{ccc|cc|cc|cc}
\toprule
\multirow{2}{*}{LLMs} & \multicolumn{4}{c}{TicTacToe}                                     & \multicolumn{4}{c}{ConnectFour}                                   \\
\cmidrule(lr){2-5}
\cmidrule(lr){6-9}

                      & Win Rate & w/o Hint & Error Rate & w/o Hint & Win Rate & w/o Hint & Error Rate & w/o Hint \\
\midrule
\midrule
GPT-4     & 69\%      & 31\%(\textcolor{red}{$\downarrow$ 38\%})               & 0\%        & 0\%                 & 31\%      & 0\%   (\textcolor{red}{$\downarrow$ 31\%})             & 25\%       & 75\% (\textcolor{red}{$\uparrow$ 50\%})               \\
GPT-3.5-Turbo         & 22\%      & 17\%(\textcolor{red}{$\downarrow$ 5\%})               & 37\%       & 38\% (\textcolor{red}{$\uparrow$ 1\%})                & 51\%      & 42\%  (\textcolor{red}{$\downarrow$ 9\%})             & 21\%       & 24\% (\textcolor{red}{$\uparrow$ 3\%})               \\
LLaMA2-70-Chat        & 56\%      & 0\%(\textcolor{red}{$\downarrow$ 56\%})                & 0\%        & 100\%  (\textcolor{red}{$\uparrow$ 100\%})             & 56\%      & 0\%   (\textcolor{red}{$\downarrow$ 56\%})             & 0\%        & 100\%    (\textcolor{red}{$\uparrow$ 100\%})           \\
Qwen-72B-Chat         & 56\%      & 0\%(\textcolor{red}{$\downarrow$ 56\%})                & 0\%        & 99\%   (\textcolor{red}{$\uparrow$ 99\%})             & 61\%      & 18\%   (\textcolor{red}{$\downarrow$ 43\%})            & 0\%        & 42\%  (\textcolor{red}{$\uparrow$ 42\%})              \\
Qwen-14B-Chat         & 56\%      & 0\% (\textcolor{red}{$\downarrow$ 56\%})               & 0\%        & 100\%   (\textcolor{red}{$\uparrow$ 100\%})            & 54\%      & 22\% (\textcolor{red}{$\downarrow$ 32\%})              & 3\%        & 63\% (\textcolor{red}{$\uparrow$ 60\%})                \\
DeepSeek-67B-Chat     & 55\%      & 2\% (\textcolor{red}{$\downarrow$ 53\%})               & 4\%        & 92\%  (\textcolor{red}{$\uparrow$ 88\%})              & 64\%      & 11\%   (\textcolor{red}{$\downarrow$ 53\%})            & 0\%        & 63\% (\textcolor{red}{$\uparrow$ 63\%})               \\
DeepSeek-7B-Chat      & 52\%      & 3\% (\textcolor{red}{$\downarrow$ 49\%})               & 7\%        & 90\%     (\textcolor{red}{$\uparrow$ 83\%})           & 36\%      & 37\% (\textcolor{teal}{$\uparrow$ 1\%})              & 4\%        & 48\%   (\textcolor{red}{$\uparrow$ 44\%})              \\
\bottomrule
\end{tabular}
}
\caption{Experimental results of seven LLMs in TicTacToe and ConnectFour environments, ``Win Rate'' represents the winning rate of LLMs with position hints in 100 games, ``Error Rate'' represents the rate of illegal moves, ``w/o Hint'' represents the results without position hint.}
\vspace{-6mm}
\label{sparsing}
\end{table*}
In this section, we will conduct a thorough analysis of the experimental results, evaluating LLMs from seven key perspectives: spatial comprehension, strategic planning, numerical reasoning, risk assessment, communication, opponent modeling, and team collaboration. We chose to experiment with 7 LLMs that demonstrate superior performance.
\paragraph{Spatial Comprehension} To investigate the spatial comprehension capabilities of LLMs, we conducted 100 self-play games with 7 different LLMs each, recording winning rates and the frequency of hallucinations (instances of illegal moves). Simultaneously, we eliminated the prompts regarding the positioning of the pieces in the LLMs' input and executed another set of 100 self-games, again tallying the winning rates and the occurrences of hallucinations. The results are shown in Table~\ref{sparsing}, which indicates that in both scenarios, the absence of positional cues significantly escalates the likelihood of LLMs experiencing hallucinations, with an average increase of 59.5\%. Consequently, this results in a substantial reduction in their win rates, which on average decline by 38.3\%.

Through a meticulous case study, we discovered that LLMs demonstrate a limited grasp of the vertical dimension on 2D chessboards. This limitation likely arises because we transform the 2D chessboard into a 1D string input for the LLMs via line breaks. Grasping the vertical concept requires the LLM to correlate two tokens separated by a specific distance in the one-dimensional input. Consequently, even though ConnectFour's chessboard is larger and its state space more complex, TicTacToe's 2D action space demands a higher level of spatial comprehension from the LLM. Hence, in the absence of positional prompts, the performance deterioration in TicTacToe is more pronounced. A case study can be found in Appendix~\ref{case:connect}
\paragraph{Strategic Planning}
\begin{figure}[b]
  \centering
  \vspace{-4mm}
  
\includegraphics[width=\linewidth]{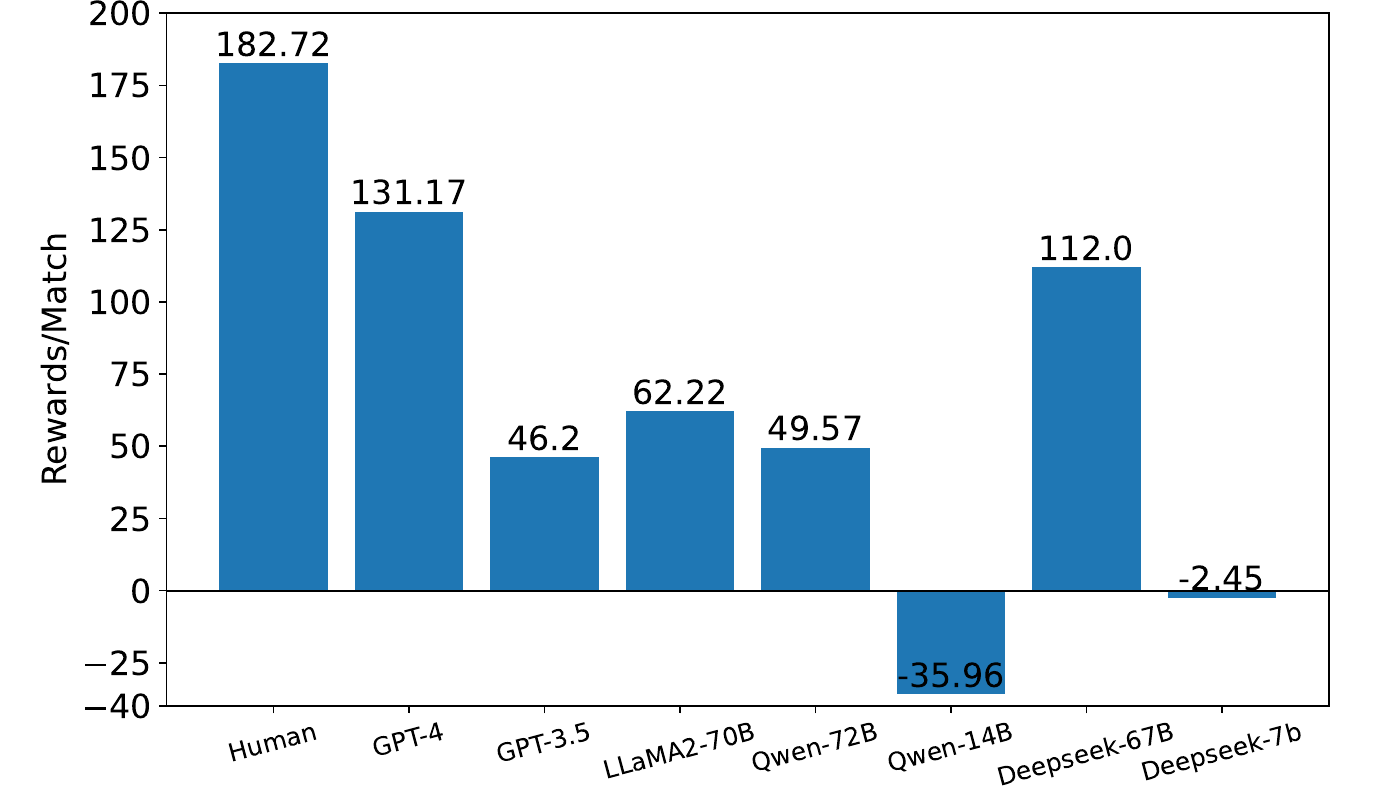}
  \caption{Average reward received by Human and 7 LLMs per game.}
  \label{figure:reward}
  \vspace{-5mm}
\end{figure}
To delve into the strategic planning capabilities of LLMs within a multi-agent environment, we crafted a specialized board valuation function tailored for the ConnectFour environment as follows:
\begin{equation}
\begin{aligned}
        V(s) &= 10 \times ( My_4(s) - Oppo_4(s) )\\ 
                &+ 5\times ( My_3(s) - Oppo_3(s) ) \\
                &+ 2\times ( My_2(s) - Oppo_2(s) ),
\end{aligned}
\end{equation}
 where $s$ represents the board status, $My_4(s)$, $My_3(s)$, and $My_2(s)$ represent the count of four, three, and two consecutive pieces of the player on the chessboard, respectively. Conversely, $Oppo_4(s)$, $Oppo_3(s)$, and $Oppo_2(s)$ represent the similar formations for the opponent's pieces
. We defined the reward as the change in board valuation between the state after the agent's move $s'$ and the state following the previous move $s$:
\begin{equation}
\begin{aligned}
        R(a,s,s') = V(s') - V(s),
\end{aligned}
\end{equation} where $a$ represents the LLM agent's action.
We conducted random battles between seven LLM agents until the rewards converged, and the results are shown in Figure~\ref{figure:reward}. It is evident that LLMs with a larger number of parameters consistently demonstrate superior strategic planning abilities. Each decision made by these LLMs propels them towards a more advantageous position, as indicated by the predominantly positive rewards they receive on average. Conversely, LLMs with a smaller parameter scale tend to earn, on average, negative rewards. This suggests an inability to accurately assess the current state of the chessboard and to formulate effective strategies, ultimately leading the game to progress in a direction that favors its opponent. To further explore the gap between LLM's strategic planning capabilities and humans, we manually played five games with seven LLMs each and human players emerged victorious in every game. Simultaneously, we observe that even GPT-4 rarely prioritizes reward maximization (such as turning a blind eye to winning positions). This tendency could stem from the fact that LLMs predominantly base their decisions on textual data. Consequently, the model must translate the game state into text descriptions before making decisions. This process of conversion may result in information loss, affecting the model's strategic performance. An error analysis can be found in Appendix~\ref{case:connect}
\paragraph{Communication}

To explore the communication capabilities of LLM, we followed the settings in Section~\ref{settings} and manually analyzed 50 game records for each LLM, calculating the percentage of their in-game clues that accurately described their secret word (``Description'' in Table~\ref{analysis_undercover}). As illustrated in Table~\ref{analysis_undercover}, LLMs demonstrated outstanding proficiency in this task. All LLMs achieved a description success rate of 95.30\% on average, with four of them reaching a perfect 100\% success rate. This high level of performance can be attributed to the LLMs being trained on extensive textual datasets, which equip them with a nuanced understanding of word associations and contextual meanings, allowing them to more easily delineate the meanings of different words. 
However, we also observed that receiving information poses a greater challenge for LLMs compared to giving information. When interpreting messages from other LLMs, they often make cognitive errors, leading to misinterpretations or distortions of the original clues. This may be because their interpretations are often based on their knowledge base rather than relying entirely on input information. This can lead them to misinterpret the intent or context of the original message, thus tampering with the original clue, which shows that LLMs still have limitations in handling complex, multi-level communications. A case study can be found in Appendix~\ref{case:undercover}

\begin{table}[]
\centering
\scalebox{0.75}{
\begin{tabular}{cccc}
\toprule
LLMs              & Description & Guess & Guess (strict)\\
\midrule
\midrule
GPT-4 & 100.00\%       & 80.00\%  & 20.00\%                          \\
GPT-3.5-Turbo     & 100.00\%       & 86.00\%  & 6.00\%                           \\
LLaMA2-70B-Chat    & 93.50\%       & 52.00\%  & 2.00\%                           \\
Qwen-72B-Chat     & 100.00\%       & 66.00\%  & 10.00\%                          \\
Qwen-14B-Chat     & 87.65\%     & 2.00\%   & 0.00\%                           \\
DeepSeek-67B-Chat & 100.00\%       & 74.00\%  & 6.00\%                          \\
DeepSeek-7B-Chat  & 85.91\%     & 38.00\%  & 0.00\%                           \\
\bottomrule
\end{tabular}}
\caption{Three different metrics obtained by 7 LLMs in the Undercover environment.}
\vspace{-8mm}
\label{analysis_undercover}
\end{table}

\paragraph{Opponent Modeling}
Moreover, the Undercover environment demands that LLMs possess not only communication skills but also the capability to model opponents. Building upon the settings established in Section~\ref{settings}, we further tasked each LLM agent with deducing the secret words of others during the voting phase. We recorded the instances where each LLM accurately guessed the secret word of any one agent (``Guess'' in Table~\ref{analysis_undercover}), as well as the occasions where they successfully identified the secret words of all other agents (``Guess (strict)'' in Table~\ref{analysis_undercover}). The results are shown in Table~\ref{analysis_undercover}, where we can find that inferring the secret words of others through their descriptions is more challenging than describing oneself. Through manual analysis, we discovered that only GPT-4 possesses the analytical and reasoning skills necessary to identify itself as an undercover agent. Subsequently, it adjusts the clues it provides in the following round for more effective camouflage. This advanced ability to model opponents also enables GPT-4 to more accurately guess the secret words of all other agents, resulting in superior performance.
\paragraph{Numerical Reasoning}
\begin{figure}[t]
  \centering
  \vspace{-4mm}
  
\includegraphics[width=\linewidth]{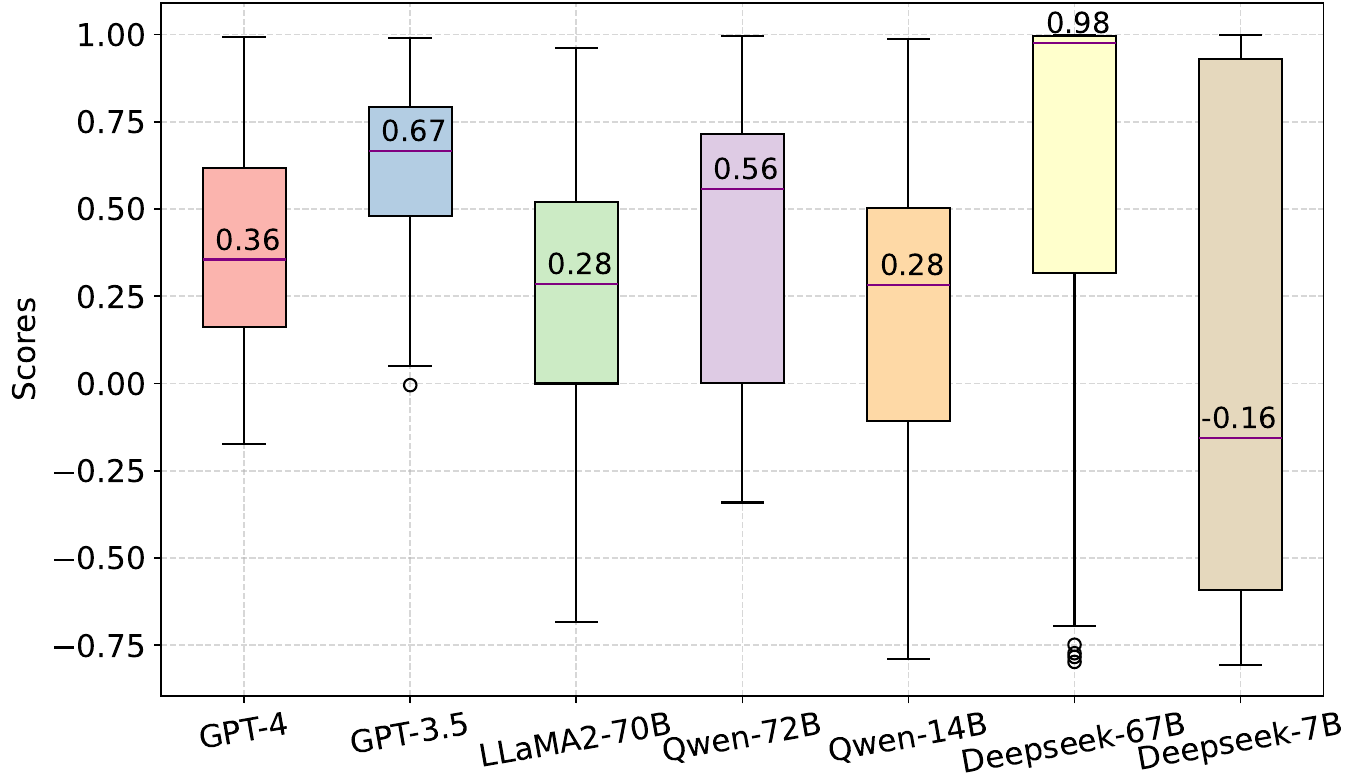}
  \caption{Scores obtained by 7 LLMs in 100 bid games.}
  \label{figure:bid}
  \vspace{-7mm}
\end{figure}
Bid represents a typical problem in incomplete information games. In such scenarios, the Bayesian Nash equilibrium for two rational agents is theoretically set at $\frac{v}{2}$, where $v$ denotes each agent's valuation of the item ~\citep{Nisan_Roughgarden_Tardos_Vazirani_2007}. We study the numerical reasoning ability of LLMs in this situation. We play 100 self-games of seven types of LLMs in the Bid environment, and count the difference between their bids $v'$ and the Nash equilibrium $\frac{v}{2}$, and calculate the normalized score $score = \frac{v'-v/2}{v/2}$. The results are shown in the Figure~\ref{figure:bid}.
The data presented in the figure allows us to deduce that nearly all LLMs, except Deepseek-7B, tend to place bids that exceed the Nash equilibrium, with their median bids being on average 52\% higher. Notably, Deepseek-67B exhibits an extreme behavior: it opts for bids that significantly sacrifice its profits, ostensibly to guarantee a win in the auction. Consequently, its overall performance in this particular environment is notably subpar as illustrated in Table~\ref{main-result}.

\begin{figure}[h]
  \centering
  \vspace{-4mm}
  
\includegraphics[width=\linewidth]{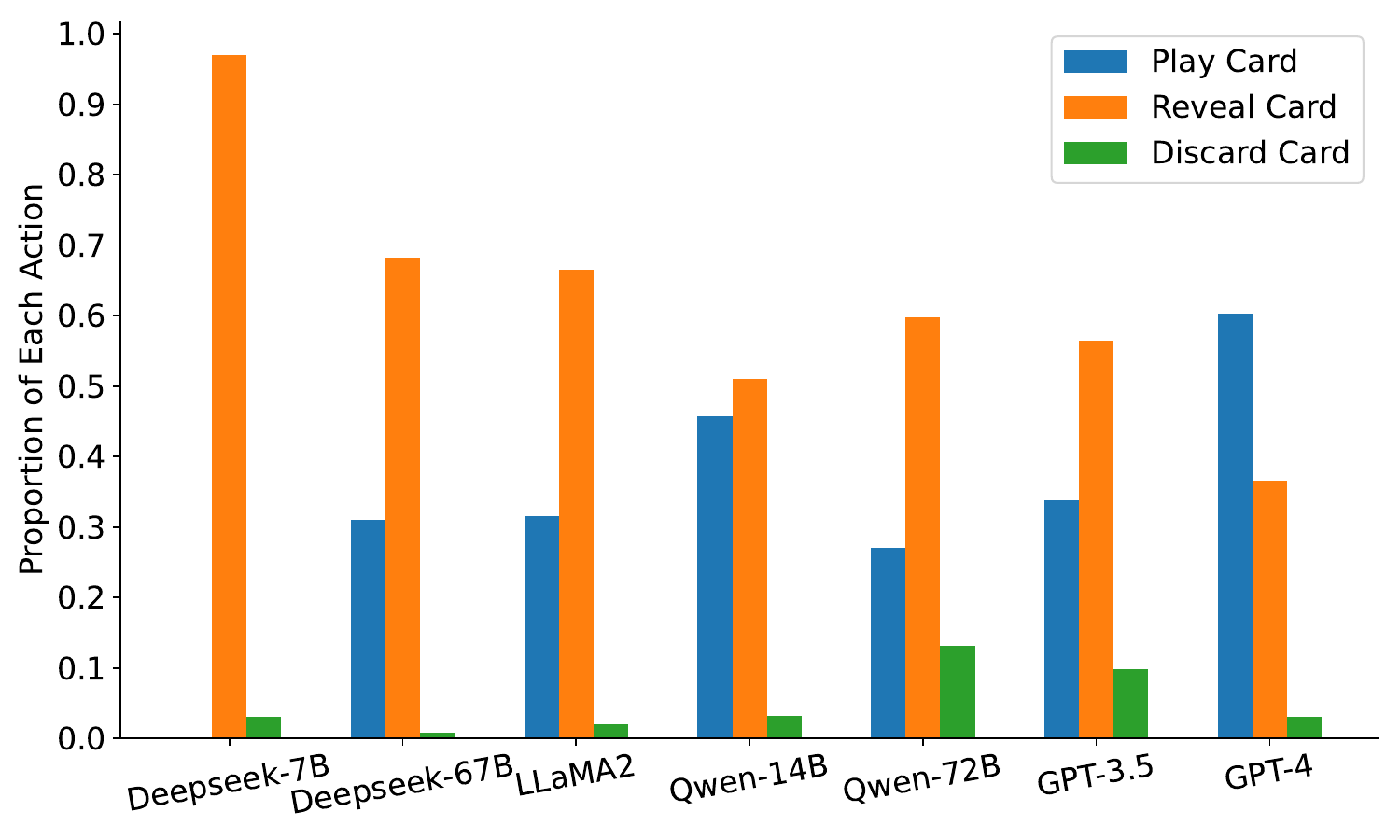}
  \caption{The proportion of three types of actions when playing Hanabi among 7 LLMs.}
  \label{figure:hanabi}
  \vspace{-5mm}
\end{figure}

\paragraph{Risk Assessment}
\begin{figure}[h]
  \centering
  \vspace{-4mm}
  
\includegraphics[width=\linewidth]{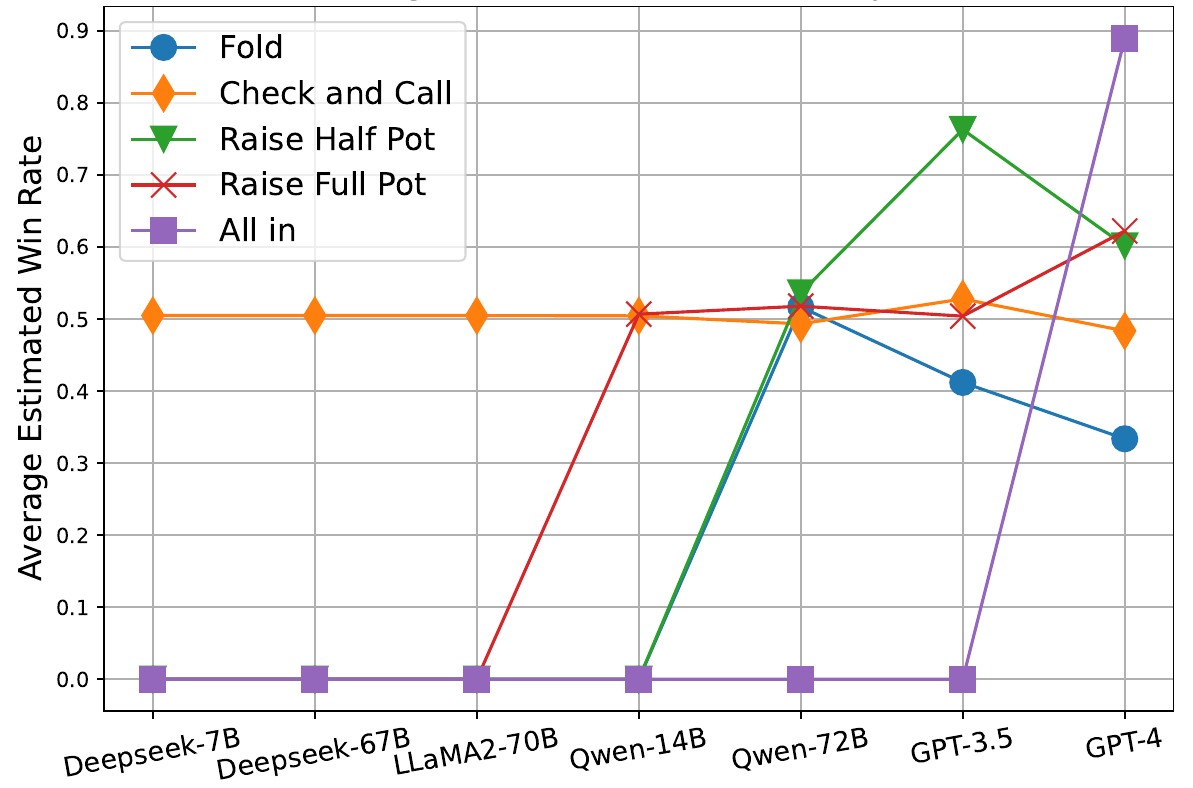}
  \caption{The average estimated win rate corresponding to each action in 100 Texas Hold'em games for 7 LLMs.}
  \label{figure:texas}
  \vspace{-5mm}
\end{figure}
To investigate the risk assessment capabilities of LLMs in an incomplete information environment, we sampled 100 static card games from the dynamic Texas Hold'em environment. In each game, the LLMs are equipped with only two private cards and three public cards as their information basis. We use Monte Carlo sampling to estimate the winning rate of LLM in each game, then record the actions taken by each LLM under these specific conditions. The specific actions of each LLM, along with the average estimated winning rate are shown in Figure~\ref{figure:texas}, where we can observe that LLMs tend to adopt a more conservative approach in their risk assessments. Additionally, three of the LLMs consistently opted for ``Check and Call'' strategies, and all models except GPT-4 refrained from selecting the most aggressive course of action, ``All in''. This may be because LLMs received caution in treating gambling-related scenarios during the pre-training and security alignment phase. Simultaneously, we noted that GPT-4 exhibits a robust risk assessment capability. It tends to choose ``All in'' when the winning rate is exceptionally high. Moreover, GPT-4 adapts its strategy as the winning rate of the hand decreases, opting for relatively low-risk actions. Notably, when the hand's winning rate falls below 40\%, GPT-4 is the sole LLM that selects ``Fold'' as a strategic decision to mitigate losses on time.
\paragraph{Team Collaboration}
As outlined in Section~\ref{settings}, within the Hanabi environment, LLMs are required to alternate between ``discarding cards'' and ``revealing cards'' to cooperate to obtain more information, and then build fireworks through ``play cards''. To investigate the efficacy of LLMs in executing team cooperation, we analyzed the frequency of each type of action performed by LLMs. As shown in Figure~\ref{figure:hanabi}, although most LLMs frequently utilize the ``reveal cards'' action to share hand information with their opponent, they seldom use the ``discard cards'' action, which would give the opponent a chance to ``reveal cards'' for more information, thus rashly choosing ``play cards'' when there is insufficient information, causing the game to fail.

\section{Related Works}
\paragraph{LLM Evaluation} With the rapid development of LLMs, the evaluation practices of traditional NLP tasks \cite{HuGCWY23, bang2023multitask,zhong2023can,qin-etal-2023-chatgpt,hu2022chef} limit their performance. Instead, the use of human-written exam questions as an assessment paradigm \cite{HuPino2023,hendryckstest2021,liang2022holistic,srivastava2023beyond} has become increasingly mainstream, focusing primarily on testing the world knowledge and reasoning capabilities in a multi-task benchmark. In addition, some newly proposed benchmarks and methods examine large language models from multiple dimensions, such as adaptability to professional fields \cite{xiang2023care}, real-world application \cite{li-etal-2023-api,liu2024survey,liu2023semantic,liu2023Unforgeable}, robustness \cite{zhu2023promptbench}, multi-modal capabilities \cite{fu2023mme}, etc. Recently, the trend of treating large language models as agents has become increasingly evident in academia, and a series of studies \cite{liu2023agentbench,wu2023smartplay} aimed to evaluate the performance of these models when acting as agents in games. However, most of these studies are limited to a single agent, thus failing to fully capture the characteristics of large language models in game or cooperative scenarios and group behavior.
\paragraph{LLM-based Agents} Reinforcement learning has been extensively employed in the training of autonomous agents \cite{ribeiro2002reinforcement,isbell2001social,mnih2013playing}. With the superior capabilities of LLMs, the utilization of LLMs as cognitive entities and controllers for agents has garnered widespread acclaim \cite{huang2022language,park2023generative,sumers2023cognitive}. \citet{wei2022chain} initially introduced the CoT technology, which notably amplifies the reasoning and planning abilities of LLM-based agents, consequently leading a research surge dominated by reasoning strategies \cite{yao2023tree,wang2023selfconsistency,zhou2023leasttomost}. Furthermore, empirical studies indicate that collaborative and adversarial frameworks \cite{li2023camel,chen2023agentverse,chan2023chateval} involving multiple LLMs surpass the capabilities of singular agents across various tasks and may lead to the emergence of social phenomena \cite{park2023generative}.
\section{Conclusion}
In our study, we introduced \modelname, a benchmark designed to assess the diverse abilities of LLM agents in dynamic, multi-agent environments. Through comprehensive analysis of seven game environments that require necessary abilities for LLM agents, our findings reveal that LLMs exhibit weaknesses in spatial reasoning, opponent modeling, and team collaboration. Enhancing the performance of LLM agents in dynamic multi-agent settings remains an unresolved challenge. We anticipate that future researchers will utilize \modelname~to conduct evaluations in a broader range of scenarios.
\section*{Limitation}
While our research primarily focuses on assessing the capabilities of LLM agents within dynamic multi-agent environments, it is important to acknowledge two limitations. Firstly, in the journey towards achieving autonomous agents, an LLM agent's ability extends beyond merely analyzing textual information. It necessitates engaging with a variety of modal inputs, including video and audio. The capabilities of LLMs within such multi-modal dynamic contexts remain an untapped area of exploration. Secondly, our study did not delve into the potential of LLMs that leverage external tools. We anticipate and encourage further research in both of these promising dimensions to broaden our understanding and capabilities of LLM agents.

\section*{Ethical Considerations}
The ethical landscape surrounding LLM agents is marked by significant challenges, particularly in the realms of responsible usage, and the potential for misuse. The deployment of autonomous LLM agents in decision-making roles raises questions about accountability and the need for robust frameworks to prevent unethical applications. The potential for misuse, such as manipulating scenarios or exploiting systems, underscores the necessity for stringent guidelines and monitoring.

\bibliography{anthology,custom}

\begin{thebibliography}{50}
\expandafter\ifx\csname natexlab\endcsname\relax\def\natexlab#1{#1}\fi

\bibitem[{Nis(2007)}]{Nisan_Roughgarden_Tardos_Vazirani_2007}
 2007.
\newblock \emph{Algorithmic Game Theory}.
\newblock Cambridge University Press.

\bibitem[{Bang et~al.(2023)Bang, Cahyawijaya, Lee, Dai, Su, Wilie, Lovenia, Ji, Yu, Chung et~al.}]{bang2023multitask}
Yejin Bang, Samuel Cahyawijaya, Nayeon Lee, Wenliang Dai, Dan Su, Bryan Wilie, Holy Lovenia, Ziwei Ji, Tiezheng Yu, Willy Chung, et~al. 2023.
\newblock A multitask, multilingual, multimodal evaluation of chatgpt on reasoning, hallucination, and interactivity.
\newblock \emph{arXiv preprint arXiv:2302.04023}.

\bibitem[{Chan et~al.(2023)Chan, Chen, Su, Yu, Xue, Zhang, Fu, and Liu}]{chan2023chateval}
Chi-Min Chan, Weize Chen, Yusheng Su, Jianxuan Yu, Wei Xue, Shanghang Zhang, Jie Fu, and Zhiyuan Liu. 2023.
\newblock Chateval: Towards better llm-based evaluators through multi-agent debate.
\newblock \emph{arXiv preprint arXiv:2308.07201}.

\bibitem[{Chen et~al.(2023)Chen, Su, Zuo, Yang, Yuan, Qian, Chan, Qin, Lu, Xie et~al.}]{chen2023agentverse}
Weize Chen, Yusheng Su, Jingwei Zuo, Cheng Yang, Chenfei Yuan, Chen Qian, Chi-Min Chan, Yujia Qin, Yaxi Lu, Ruobing Xie, et~al. 2023.
\newblock Agentverse: Facilitating multi-agent collaboration and exploring emergent behaviors in agents.
\newblock \emph{arXiv preprint arXiv:2308.10848}.

\bibitem[{Foster et~al.(2023)Foster, Foster, Golowich, and Rakhlin}]{foster2023complexity}
Dean Foster, Dylan~J Foster, Noah Golowich, and Alexander Rakhlin. 2023.
\newblock On the complexity of multi-agent decision making: From learning in games to partial monitoring.
\newblock In \emph{The Thirty Sixth Annual Conference on Learning Theory}, pages 2678--2792. PMLR.

\bibitem[{Fu et~al.(2023)Fu, Chen, Shen, Qin, Zhang, Lin, Yang, Zheng, Li, Sun et~al.}]{fu2023mme}
Chaoyou Fu, Peixian Chen, Yunhang Shen, Yulei Qin, Mengdan Zhang, Xu~Lin, Jinrui Yang, Xiawu Zheng, Ke~Li, Xing Sun, et~al. 2023.
\newblock Mme: A comprehensive evaluation benchmark for multimodal large language models.
\newblock \emph{arXiv preprint arXiv:2306.13394}.

\bibitem[{Guo et~al.(2023)Guo, Li, Li, Tiong, Li, Tao, and Hoi}]{guo2023images}
Jiaxian Guo, Junnan Li, Dongxu Li, Anthony Meng~Huat Tiong, Boyang Li, Dacheng Tao, and Steven Hoi. 2023.
\newblock From images to textual prompts: Zero-shot visual question answering with frozen large language models.
\newblock In \emph{Proceedings of the IEEE/CVF Conference on Computer Vision and Pattern Recognition}, pages 10867--10877.

\bibitem[{Hendrycks et~al.(2021)Hendrycks, Burns, Basart, Zou, Mazeika, Song, and Steinhardt}]{hendryckstest2021}
Dan Hendrycks, Collin Burns, Steven Basart, Andy Zou, Mantas Mazeika, Dawn Song, and Jacob Steinhardt. 2021.
\newblock Measuring massive multitask language understanding.
\newblock \emph{Proceedings of the International Conference on Learning Representations (ICLR)}.

\bibitem[{Herbrich et~al.(2006)Herbrich, Minka, and Graepel}]{herbrich2006trueskill}
Ralf Herbrich, Tom Minka, and Thore Graepel. 2006.
\newblock Trueskill™: a bayesian skill rating system.
\newblock \emph{Advances in neural information processing systems}, 19.

\bibitem[{Hu et~al.(2024)Hu, Chen, Li, Guo, Wen, Yu, and Guo}]{HuPino2023}
Xuming Hu, Junzhe Chen, Xiaochuan Li, Yufei Guo, Lijie Wen, Philip~S. Yu, and Zhijiang Guo. 2024.
\newblock \href {https://doi.org/10.48550/ARXIV.2310.05177} {Do large language models know about facts?}
\newblock \emph{12th International Conference on Learning Representations, {ICLR} 2024, Messe Wien Exhibition and Congress Center, Vienna Austria May 7th, 2024 to May 11th, 2024}, 2024.

\bibitem[{Hu et~al.(2023)Hu, Guo, Chen, Wen, and Yu}]{HuGCWY23}
Xuming Hu, Zhijiang Guo, Junzhe Chen, Lijie Wen, and Philip~S. Yu. 2023.
\newblock \href {https://doi.org/10.1145/3539618.3591896} {{MR2:} {A} benchmark for multimodal retrieval-augmented rumor detection in social media}.
\newblock In \emph{Proceedings of the 46th International {ACM} {SIGIR} Conference on Research and Development in Information Retrieval, {SIGIR} 2023, Taipei, Taiwan, July 23-27, 2023}, pages 2901--2912. {ACM}.

\bibitem[{Hu et~al.(2022)Hu, Guo, Wu, Liu, Wen, and Yu}]{hu2022chef}
Xuming Hu, Zhijiang Guo, Guanyu Wu, Aiwei Liu, Lijie Wen, and Philip~S Yu. 2022.
\newblock Chef: A pilot chinese dataset for evidence-based fact-checking.
\newblock \emph{arXiv preprint arXiv:2206.11863}.

\bibitem[{Huang et~al.(2022)Huang, Abbeel, Pathak, and Mordatch}]{huang2022language}
Wenlong Huang, Pieter Abbeel, Deepak Pathak, and Igor Mordatch. 2022.
\newblock Language models as zero-shot planners: Extracting actionable knowledge for embodied agents.
\newblock In \emph{International Conference on Machine Learning}, pages 9118--9147. PMLR.

\bibitem[{Isbell et~al.(2001)Isbell, Shelton, Kearns, Singh, and Stone}]{isbell2001social}
Charles Isbell, Christian~R Shelton, Michael Kearns, Satinder Singh, and Peter Stone. 2001.
\newblock A social reinforcement learning agent.
\newblock In \emph{Proceedings of the fifth international conference on Autonomous agents}, pages 377--384.

\bibitem[{Kojima et~al.(2022)Kojima, Gu, Reid, Matsuo, and Iwasawa}]{kojima2022large}
Takeshi Kojima, Shixiang~Shane Gu, Machel Reid, Yutaka Matsuo, and Yusuke Iwasawa. 2022.
\newblock Large language models are zero-shot reasoners.
\newblock \emph{Advances in neural information processing systems}, 35:22199--22213.

\bibitem[{Lewis et~al.(2017)Lewis, Yarats, Dauphin, Parikh, and Batra}]{lewis2017deal}
Mike Lewis, Denis Yarats, Yann~N Dauphin, Devi Parikh, and Dhruv Batra. 2017.
\newblock Deal or no deal? end-to-end learning for negotiation dialogues.
\newblock \emph{arXiv preprint arXiv:1706.05125}.

\bibitem[{Li et~al.(2023{\natexlab{a}})Li, Hammoud, Itani, Khizbullin, and Ghanem}]{li2023camel}
Guohao Li, Hasan Abed Al~Kader Hammoud, Hani Itani, Dmitrii Khizbullin, and Bernard Ghanem. 2023{\natexlab{a}}.
\newblock \href {https://openreview.net/forum?id=3IyL2XWDkG} {{CAMEL}: Communicative agents for ''mind'' exploration of large language model society}.
\newblock In \emph{Thirty-seventh Conference on Neural Information Processing Systems}.

\bibitem[{Li et~al.(2023{\natexlab{b}})Li, Zhao, Yu, Song, Li, Yu, Li, Huang, and Li}]{li-etal-2023-api}
Minghao Li, Yingxiu Zhao, Bowen Yu, Feifan Song, Hangyu Li, Haiyang Yu, Zhoujun Li, Fei Huang, and Yongbin Li. 2023{\natexlab{b}}.
\newblock \href {https://doi.org/10.18653/v1/2023.emnlp-main.187} {{API}-bank: A comprehensive benchmark for tool-augmented {LLM}s}.
\newblock In \emph{Proceedings of the 2023 Conference on Empirical Methods in Natural Language Processing}, pages 3102--3116, Singapore. Association for Computational Linguistics.

\bibitem[{Liang et~al.(2022)Liang, Bommasani, Lee, Tsipras, Soylu, Yasunaga, Zhang, Narayanan, Wu, Kumar et~al.}]{liang2022holistic}
Percy Liang, Rishi Bommasani, Tony Lee, Dimitris Tsipras, Dilara Soylu, Michihiro Yasunaga, Yian Zhang, Deepak Narayanan, Yuhuai Wu, Ananya Kumar, et~al. 2022.
\newblock Holistic evaluation of language models.
\newblock \emph{arXiv preprint arXiv:2211.09110}.

\bibitem[{Liu et~al.(2024{\natexlab{a}})Liu, Pan, Hu, Li, Wen, King, and Yu}]{liu2023Unforgeable}
Aiwei Liu, Leyi Pan, Xuming Hu, Shu'ang Li, Lijie Wen, Irwin King, and Philip~S. Yu. 2024{\natexlab{a}}.
\newblock \href {https://doi.org/10.48550/ICLR.2307.16230} {An unforgeable publicly verifiable watermark for large language models}.
\newblock \emph{12th International Conference on Learning Representations, {ICLR} 2024, Messe Wien Exhibition and Congress Center, Vienna Austria May 7th, 2024 to May 11th, 2024}, 2024.

\bibitem[{Liu et~al.(2024{\natexlab{b}})Liu, Pan, Hu, Meng, and Wen}]{liu2023semantic}
Aiwei Liu, Leyi Pan, Xuming Hu, Shiao Meng, and Lijie Wen. 2024{\natexlab{b}}.
\newblock \href {https://doi.org/10.48550/ICLR.2310.06356} {A semantic invariant robust watermark for large language models}.
\newblock \emph{12th International Conference on Learning Representations, {ICLR} 2024, Messe Wien Exhibition and Congress Center, Vienna Austria May 7th, 2024 to May 11th, 2024}, 2024.

\bibitem[{Liu et~al.(2024{\natexlab{c}})Liu, Pan, Lu, Li, Hu, Zhang, Wen, King, Xiong, and Yu}]{liu2024survey}
Aiwei Liu, Leyi Pan, Yijian Lu, Jingjing Li, Xuming Hu, Xi~Zhang, Lijie Wen, Irwin King, Hui Xiong, and Philip~S. Yu. 2024{\natexlab{c}}.
\newblock \href {http://arxiv.org/abs/2312.07913} {A survey of text watermarking in the era of large language models}.

\bibitem[{Liu et~al.(2023)Liu, Yu, Zhang, Xu, Lei, Lai, Gu, Ding, Men, Yang et~al.}]{liu2023agentbench}
Xiao Liu, Hao Yu, Hanchen Zhang, Yifan Xu, Xuanyu Lei, Hanyu Lai, Yu~Gu, Hangliang Ding, Kaiwen Men, Kejuan Yang, et~al. 2023.
\newblock Agentbench: Evaluating llms as agents.
\newblock \emph{arXiv preprint arXiv:2308.03688}.

\bibitem[{Mnih et~al.(2013)Mnih, Kavukcuoglu, Silver, Graves, Antonoglou, Wierstra, and Riedmiller}]{mnih2013playing}
Volodymyr Mnih, Koray Kavukcuoglu, David Silver, Alex Graves, Ioannis Antonoglou, Daan Wierstra, and Martin Riedmiller. 2013.
\newblock Playing atari with deep reinforcement learning.
\newblock \emph{arXiv preprint arXiv:1312.5602}.

\bibitem[{OpenAI(2022)}]{chatgpt}
OpenAI. 2022.
\newblock \href {https://chat.openai.com} {Chat{GPT}}.

\bibitem[{OpenAI(2023)}]{gpt4}
OpenAI. 2023.
\newblock \href {https://doi.org/10.48550/arXiv.2303.08774} {{GPT-4} {T}echnical {R}eport}.
\newblock \emph{CoRR}, abs/2303.08774.

\bibitem[{Pan et~al.(2024)Pan, Luo, Wang, Chen, Wang, and Wu}]{pan2024unifying}
Shirui Pan, Linhao Luo, Yufei Wang, Chen Chen, Jiapu Wang, and Xindong Wu. 2024.
\newblock Unifying large language models and knowledge graphs: A roadmap.
\newblock \emph{IEEE Transactions on Knowledge and Data Engineering}.

\bibitem[{Park et~al.(2023)Park, O'Brien, Cai, Morris, Liang, and Bernstein}]{park2023generative}
Joon~Sung Park, Joseph O'Brien, Carrie~Jun Cai, Meredith~Ringel Morris, Percy Liang, and Michael~S Bernstein. 2023.
\newblock Generative agents: Interactive simulacra of human behavior.
\newblock In \emph{Proceedings of the 36th Annual ACM Symposium on User Interface Software and Technology}, pages 1--22.

\bibitem[{Qian et~al.(2023)Qian, Cong, Yang, Chen, Su, Xu, Liu, and Sun}]{qian2023communicative}
Chen Qian, Xin Cong, Cheng Yang, Weize Chen, Yusheng Su, Juyuan Xu, Zhiyuan Liu, and Maosong Sun. 2023.
\newblock Communicative agents for software development.
\newblock \emph{arXiv preprint arXiv:2307.07924}.

\bibitem[{Qin et~al.(2023)Qin, Zhang, Zhang, Chen, Yasunaga, and Yang}]{qin-etal-2023-chatgpt}
Chengwei Qin, Aston Zhang, Zhuosheng Zhang, Jiaao Chen, Michihiro Yasunaga, and Diyi Yang. 2023.
\newblock \href {https://doi.org/10.18653/v1/2023.emnlp-main.85} {Is {C}hat{GPT} a general-purpose natural language processing task solver?}
\newblock In \emph{Proceedings of the 2023 Conference on Empirical Methods in Natural Language Processing}, pages 1339--1384, Singapore. Association for Computational Linguistics.

\bibitem[{Ribeiro(2002)}]{ribeiro2002reinforcement}
CHCR Ribeiro. 2002.
\newblock Reinforcement learning agents.
\newblock \emph{Artificial intelligence review}, 17:223--250.

\bibitem[{Srivastava et~al.(2023)Srivastava, Rastogi, Rao, Shoeb, Abid, Fisch, Brown, Santoro, Gupta, Garriga-Alonso et~al.}]{srivastava2023beyond}
Aarohi Srivastava, Abhinav Rastogi, Abhishek Rao, Abu Awal~Md Shoeb, Abubakar Abid, Adam Fisch, Adam~R Brown, Adam Santoro, Aditya Gupta, Adri{\`a} Garriga-Alonso, et~al. 2023.
\newblock \href {https://openreview.net/forum?id=uyTL5Bvosj} {Beyond the imitation game: Quantifying and extrapolating the capabilities of language models}.
\newblock \emph{Transactions on Machine Learning Research}.

\bibitem[{Sumers et~al.(2023)Sumers, Yao, Narasimhan, and Griffiths}]{sumers2023cognitive}
Theodore~R Sumers, Shunyu Yao, Karthik Narasimhan, and Thomas~L Griffiths. 2023.
\newblock Cognitive architectures for language agents.
\newblock \emph{arXiv preprint arXiv:2309.02427}.

\bibitem[{Terry et~al.(2021)Terry, Black, Grammel, Jayakumar, Hari, Sullivan, Santos, Dieffendahl, Horsch, Perez-Vicente et~al.}]{terry2021pettingzoo}
J~Terry, Benjamin Black, Nathaniel Grammel, Mario Jayakumar, Ananth Hari, Ryan Sullivan, Luis~S Santos, Clemens Dieffendahl, Caroline Horsch, Rodrigo Perez-Vicente, et~al. 2021.
\newblock Pettingzoo: Gym for multi-agent reinforcement learning.
\newblock \emph{Advances in Neural Information Processing Systems}, 34:15032--15043.

\bibitem[{Touvron et~al.(2023{\natexlab{a}})Touvron, Lavril, Izacard, Martinet, Lachaux, Lacroix, Rozi{\`{e}}re, Goyal, Hambro, Azhar, Rodriguez, Joulin, Grave, and Lample}]{llama}
Hugo Touvron, Thibaut Lavril, Gautier Izacard, Xavier Martinet, Marie{-}Anne Lachaux, Timoth{\'{e}}e Lacroix, Baptiste Rozi{\`{e}}re, Naman Goyal, Eric Hambro, Faisal Azhar, Aur{\'{e}}lien Rodriguez, Armand Joulin, Edouard Grave, and Guillaume Lample. 2023{\natexlab{a}}.
\newblock \href {https://doi.org/10.48550/arXiv.2302.13971} {Llama: Open and efficient foundation language models}.
\newblock \emph{CoRR}, abs/2302.13971.

\bibitem[{Touvron et~al.(2023{\natexlab{b}})Touvron, Martin, Stone, Albert, Almahairi, Babaei, Bashlykov, Batra, Bhargava, Bhosale, Bikel, Blecher, Canton{-}Ferrer, Chen, Cucurull, Esiobu, Fernandes, Fu, Fu, Fuller, Gao, Goswami, Goyal, Hartshorn, Hosseini, Hou, Inan, Kardas, Kerkez, Khabsa, Kloumann, Korenev, Koura, Lachaux, Lavril, Lee, Liskovich, Lu, Mao, Martinet, Mihaylov, Mishra, Molybog, Nie, Poulton, Reizenstein, Rungta, Saladi, Schelten, Silva, Smith, Subramanian, Tan, Tang, Taylor, Williams, Kuan, Xu, Yan, Zarov, Zhang, Fan, Kambadur, Narang, Rodriguez, Stojnic, Edunov, and Scialom}]{llama2}
Hugo Touvron, Louis Martin, Kevin Stone, Peter Albert, Amjad Almahairi, Yasmine Babaei, Nikolay Bashlykov, Soumya Batra, Prajjwal Bhargava, Shruti Bhosale, Dan Bikel, Lukas Blecher, Cristian Canton{-}Ferrer, Moya Chen, Guillem Cucurull, David Esiobu, Jude Fernandes, Jeremy Fu, Wenyin Fu, Brian Fuller, Cynthia Gao, Vedanuj Goswami, Naman Goyal, Anthony Hartshorn, Saghar Hosseini, Rui Hou, Hakan Inan, Marcin Kardas, Viktor Kerkez, Madian Khabsa, Isabel Kloumann, Artem Korenev, Punit~Singh Koura, Marie{-}Anne Lachaux, Thibaut Lavril, Jenya Lee, Diana Liskovich, Yinghai Lu, Yuning Mao, Xavier Martinet, Todor Mihaylov, Pushkar Mishra, Igor Molybog, Yixin Nie, Andrew Poulton, Jeremy Reizenstein, Rashi Rungta, Kalyan Saladi, Alan Schelten, Ruan Silva, Eric~Michael Smith, Ranjan Subramanian, Xiaoqing~Ellen Tan, Binh Tang, Ross Taylor, Adina Williams, Jian~Xiang Kuan, Puxin Xu, Zheng Yan, Iliyan Zarov, Yuchen Zhang, Angela Fan, Melanie Kambadur, Sharan Narang, Aur{\'{e}}lien Rodriguez, Robert Stojnic, Sergey Edunov,
  and Thomas Scialom. 2023{\natexlab{b}}.
\newblock \href {https://doi.org/10.48550/arXiv.2307.09288} {Llama 2: Open foundation and fine-tuned chat models}.
\newblock \emph{CoRR}, abs/2307.09288.

\bibitem[{Wang et~al.(2023{\natexlab{a}})Wang, Li, Chen, Zhu, Lin, Cao, Liu, Liu, and Sui}]{wang2023large}
Peiyi Wang, Lei Li, Liang Chen, Dawei Zhu, Binghuai Lin, Yunbo Cao, Qi~Liu, Tianyu Liu, and Zhifang Sui. 2023{\natexlab{a}}.
\newblock Large language models are not fair evaluators.
\newblock \emph{arXiv preprint arXiv:2305.17926}.

\bibitem[{Wang et~al.(2023{\natexlab{b}})Wang, Wei, Schuurmans, Le, Chi, Narang, Chowdhery, and Zhou}]{wang2023selfconsistency}
Xuezhi Wang, Jason Wei, Dale Schuurmans, Quoc~V Le, Ed~H. Chi, Sharan Narang, Aakanksha Chowdhery, and Denny Zhou. 2023{\natexlab{b}}.
\newblock \href {https://openreview.net/forum?id=1PL1NIMMrw} {Self-consistency improves chain of thought reasoning in language models}.
\newblock In \emph{The Eleventh International Conference on Learning Representations}.

\bibitem[{Wei et~al.(2022)Wei, Wang, Schuurmans, Bosma, Xia, Chi, Le, Zhou et~al.}]{wei2022chain}
Jason Wei, Xuezhi Wang, Dale Schuurmans, Maarten Bosma, Fei Xia, Ed~Chi, Quoc~V Le, Denny Zhou, et~al. 2022.
\newblock Chain-of-thought prompting elicits reasoning in large language models.
\newblock \emph{Advances in Neural Information Processing Systems}, 35:24824--24837.

\bibitem[{Wei et~al.(2023)Wei, Cui, Cheng, Wang, Zhang, Huang, Xie, Xu, Chen, Zhang et~al.}]{wei2023zero}
Xiang Wei, Xingyu Cui, Ning Cheng, Xiaobin Wang, Xin Zhang, Shen Huang, Pengjun Xie, Jinan Xu, Yufeng Chen, Meishan Zhang, et~al. 2023.
\newblock Zero-shot information extraction via chatting with chatgpt.
\newblock \emph{arXiv preprint arXiv:2302.10205}.

\bibitem[{Wu et~al.(2023)Wu, Tang, Mitchell, and Li}]{wu2023smartplay}
Yue Wu, Xuan Tang, Tom~M Mitchell, and Yuanzhi Li. 2023.
\newblock Smartplay: A benchmark for llms as intelligent agents.
\newblock \emph{arXiv preprint arXiv:2310.01557}.

\bibitem[{Xiang et~al.(2023)Xiang, Li, Li, Bai, Wei, Wang, and Garcia}]{xiang2023care}
Tong Xiang, Liangzhi Li, Wangyue Li, Mingbai Bai, Lu~Wei, Bowen Wang, and Noa Garcia. 2023.
\newblock Care-mi: chinese benchmark for misinformation evaluation in maternity and infant care.
\newblock \emph{arXiv preprint arXiv:2307.01458}.

\bibitem[{Xu et~al.(2023)Xu, Hu, Zhou, Ren, Dong, Keutzer, Ng, and Feng}]{xu2023magic}
Lin Xu, Zhiyuan Hu, Daquan Zhou, Hongyu Ren, Zhen Dong, Kurt Keutzer, See~Kiong Ng, and Jiashi Feng. 2023.
\newblock Magic: Benchmarking large language model powered multi-agent in cognition, adaptability, rationality and collaboration.
\newblock \emph{arXiv preprint arXiv:2311.08562}.

\bibitem[{Yang et~al.(2023)Yang, Jin, Tang, Han, Feng, Jiang, Yin, and Hu}]{yang2023harnessing}
Jingfeng Yang, Hongye Jin, Ruixiang Tang, Xiaotian Han, Qizhang Feng, Haoming Jiang, Bing Yin, and Xia Hu. 2023.
\newblock Harnessing the power of llms in practice: A survey on chatgpt and beyond.
\newblock \emph{arXiv preprint arXiv:2304.13712}.

\bibitem[{Yao et~al.(2022)Yao, Chen, Yang, and Narasimhan}]{yao2022webshop}
Shunyu Yao, Howard Chen, John Yang, and Karthik Narasimhan. 2022.
\newblock Webshop: Towards scalable real-world web interaction with grounded language agents.
\newblock \emph{Advances in Neural Information Processing Systems}, 35:20744--20757.

\bibitem[{Yao et~al.(2023)Yao, Yu, Zhao, Shafran, Griffiths, Cao, and Narasimhan}]{yao2023tree}
Shunyu Yao, Dian Yu, Jeffrey Zhao, Izhak Shafran, Thomas~L. Griffiths, Yuan Cao, and Karthik~R Narasimhan. 2023.
\newblock \href {https://openreview.net/forum?id=5Xc1ecxO1h} {Tree of thoughts: Deliberate problem solving with large language models}.
\newblock In \emph{Thirty-seventh Conference on Neural Information Processing Systems}.

\bibitem[{Zhong et~al.(2023)Zhong, Ding, Liu, Du, and Tao}]{zhong2023can}
Qihuang Zhong, Liang Ding, Juhua Liu, Bo~Du, and Dacheng Tao. 2023.
\newblock Can chatgpt understand too? a comparative study on chatgpt and fine-tuned bert.
\newblock \emph{arXiv preprint arXiv:2302.10198}.

\bibitem[{Zhou et~al.(2023{\natexlab{a}})Zhou, Sch{\"a}rli, Hou, Wei, Scales, Wang, Schuurmans, Cui, Bousquet, Le, and Chi}]{zhou2023leasttomost}
Denny Zhou, Nathanael Sch{\"a}rli, Le~Hou, Jason Wei, Nathan Scales, Xuezhi Wang, Dale Schuurmans, Claire Cui, Olivier Bousquet, Quoc~V Le, and Ed~H. Chi. 2023{\natexlab{a}}.
\newblock \href {https://openreview.net/forum?id=WZH7099tgfM} {Least-to-most prompting enables complex reasoning in large language models}.
\newblock In \emph{The Eleventh International Conference on Learning Representations}.

\bibitem[{Zhou et~al.(2023{\natexlab{b}})Zhou, Zhu, Chen, Chen, Zhao, Chen, Lin, Wen, and Han}]{zhou2023don}
Kun Zhou, Yutao Zhu, Zhipeng Chen, Wentong Chen, Wayne~Xin Zhao, Xu~Chen, Yankai Lin, Ji-Rong Wen, and Jiawei Han. 2023{\natexlab{b}}.
\newblock Don't make your llm an evaluation benchmark cheater.
\newblock \emph{arXiv preprint arXiv:2311.01964}.

\bibitem[{Zhu et~al.(2023)Zhu, Wang, Zhou, Wang, Chen, Wang, Yang, Ye, Gong, Zhang et~al.}]{zhu2023promptbench}
Kaijie Zhu, Jindong Wang, Jiaheng Zhou, Zichen Wang, Hao Chen, Yidong Wang, Linyi Yang, Wei Ye, Neil~Zhenqiang Gong, Yue Zhang, et~al. 2023.
\newblock Promptbench: Towards evaluating the robustness of large language models on adversarial prompts.
\newblock \emph{arXiv preprint arXiv:2306.04528}.

\end{thebibliography}
\bibliographystyle{acl_natbib}
\onecolumn
\newpage

\appendix

\newtcolorbox{promptbox}[1][]{
  breakable,
  title=#1,
  colback=gray!5,
  colframe=black,
  colbacktitle=gray!15,
  coltitle=black,
  fonttitle=\bfseries,
  bottomrule=1.5pt,
  toprule=1.5pt,
  leftrule=1pt,
  rightrule=1pt,
  arc=0pt,
  outer arc=0pt,
  enhanced,
  before upper={\parindent=1.5em} 
}

\newcommand{\fourconnect}[1]
    {
    \begin{tikzpicture}[line width=2pt]
        \def\r{3mm}
            \tikzset{
                    circ/.pic={\draw circle (\r);},
                    cross/.pic={\draw (-\r,-\r) -- (\r,\r) (-\r,\r) -- (\r,-\r);},
                    opt/.pic={\draw[opacity=0.2] (-\r,-\r) -- (\r,\r) (-\r,\r) -- (\r,-\r);}
                    }
            
            \foreach \i in {1,2,3} \draw (\i,0) -- (\i,4) (0,\i) -- (4,\i) ;
            
            \setcounter{num}{0}
            \foreach \y in {0,...,3}
                \foreach \x in {0,...,3}
                    {
                    \coordinate (\thenum) at (\x+0.5,3-\y+0.5);
                    \addtocounter{num}{1}
                    }

        \def\X{X} \def\x{x} \def\O{O} \def\n{n}
        
        \foreach \l [count = \i from 0] in {#1}
            {
            \if\l\X \path (\i) pic{cross};
            \else
                \if\l\O \path (\i) pic{circ};
                \else
                    \if\l\x \path (\i) pic{opt};
                    \else
                        \if\l\n \node[opacity=0.5] at (\i) {\sffamily\i};
                        \fi
                    \fi
                \fi
            \fi
            }
    \end{tikzpicture}
    }
\section{Complete Prompt Design}
\label{appendix:prompt}

In this section, we provide complete prompts for all seven game environments in LLM Arena. As the agent in the LLM Arena must consistently interact with the environment, similar to the reinforcement learning setting, various prompts are required to guide the agent in comprehending the rules and taking action. Specifically, three types of prompts work:

\begin{itemize}
  \item \textbf{System Prompt.} System prompts are special messages used to steer the behavior of LLM. They allow to prescribe the agent's style and task within certain bounds, making it more customizable and adaptable for various use cases.
  
  In LLM Arena, we stipulate the role of the agent and the rules of the game in the system prompt.
  \item \textbf{Observation Prompt.} Observation Prompt provides the necessary state information for the agent to interact with the environment, including the opponent's actions, current game status, legal actions, etc. We use \{\} to frame these changing information.
  \item \textbf{Action Prompt.} Action Prompt guides the Agent to choose from legal actions and output regularized actions. In addition, to stimulate the agent's reasoning ability, in some environments, we also add CoT Prompt.
\end{itemize}

\subsection{TicTacToe}

\newcounter{num}
\newcommand{\tictactoe}[1]
    {
    \begin{tikzpicture}[line width=2pt]
        \def\r{3mm}
            \tikzset{
                    circ/.pic={\draw circle (\r);},
                    cross/.pic={\draw (-\r,-\r) -- (\r,\r) (-\r,\r) -- (\r,-\r);},
                    opt/.pic={\draw[opacity=0.2] (-\r,-\r) -- (\r,\r) (-\r,\r) -- (\r,-\r);}
                    }
            
            \foreach \i in {1,2} \draw (\i,0) -- (\i,3) (0,\i) -- (3,\i);
            
            \setcounter{num}{0}
            \foreach \y in {0,...,2}
                \foreach \x in {0,...,2}
                    {
                    \coordinate (\thenum) at (\x+0.5,2-\y+0.5);
                    \addtocounter{num}{1}
                    }

        \def\X{X} \def\x{x} \def\O{O} \def\n{n}
        
        \foreach \l [count = \i from 0] in {#1}
            {
            \if\l\X \path (\i) pic{cross};
            \else
                \if\l\O \path (\i) pic{circ};
                \else
                    \if\l\x \path (\i) pic{opt};
                    \else
                        \if\l\n \node[opacity=0.5] at (\i) {\sffamily\i};
                        \fi
                    \fi
                \fi
            \fi
            }
    \end{tikzpicture}
    }

In TicTacToe, the content in \{\} can be obtained from the interaction between the agent and the environment. For example, \{player\_type\} includes "X" and "O", \{board\_status\} represents the current chessboard, and we describe the entire chessboard in the form of text, as the example in System Prompt. \{available\} represents the current legal action list of the agent.

The detailed prompt template of TicTacToe is as follows.

\newcommand\VRule{\vrule width 1pt}
\begin{promptbox}[System Prompt]
You are playing TicTacToe. 

TicTacToe is played on a three-by-three grid by two players, who alternately place the marks X and O in one of the nine spaces in the grid. 

The player who succeeds in placing three of their marks in a horizontal, vertical, or diagonal row is the winner.

In the following example, the first player (X) wins the game in seven steps:

1. [Player 1]: X: (1, 3)

\begin{center}
\tictactoe{  ,,X,
            ,, ,
            , ,}    
\end{center}


2. [Player 2]: O: (1, 1)

\begin{center}
\tictactoe{  O,,X,
            ,, ,
            , ,}   
\end{center}


3. [Player 1]: X: (3, 1)

\begin{center}
\tictactoe{  O,,X,
            ,, ,
            X, ,}     
\end{center}


4.  [Player 2]: O: (2, 2)

\begin{center}
\tictactoe{  O,,X,
            ,O, ,
            X, ,}  
\end{center}

5. [Player 1]: X: (3, 3)

\begin{center}
\tictactoe{  O,,X,
            ,O, ,
            X, ,X}  
\end{center}


6. [Player 2]: O: (2, 3)

\begin{center}
\tictactoe{  O,,X,
            ,O, O,
            X, ,X}  
\end{center}


7. [Player 1]: X: (3, 2)

\begin{center}
\tictactoe{  O,,X,
            ,O, O,
            X,X ,X}   
\end{center}


X plays first. Players will specify the position of the stone and the moderator will plot the board status.

If a position has been marked, future marks cannot be put in the same position.

The players interact with the game by specifying the position of the stones (x, y), where x indicates the row and y indicates the column, so (1, 1) is the top left corner and (3, 3) is the bottom right corner."
\end{promptbox}
\begin{promptbox}[Observation Prompt]
You play \{player\_type\}.

The board status is \{board\_status\}.

You can only put the mark on [\{available\}].
\end{promptbox}
\begin{promptbox}[Action Prompt]
You should only output \{player\_type\} and the position of the move, for example: "\{player\_type\}: (1, 3)"

The position you put the mark on must be empty. 

Don't say anything besides mark position.
\end{promptbox}

\subsection{ConnectFour}
In ConnectFour, the observation prompt and action prompt are similar to TicTacToe. We also use text to describe the chessboard.

The detailed prompt template of Texas Hold’em is as follows.

\begin{promptbox}[System Prompt]
You are playing ConnectFour. 

ConnectFour is played on a six-by-seven grid by two players, who alternately drop their makers X and O into one of the seven columns. Each marker will fall to the lowest available space within the selected column. 

The player who succeeds in placing four of their makers in a horizontal, vertical, or diagonal row is the winner.
X plays first. Players interact with the game by specifying the column number where they want to drop their marker. If a column is already full, players cannot drop a marker into that column.

The columns are numbered from 1 to 7, from left to right. Players cannot place a token outside this range.
\end{promptbox}
\begin{promptbox}[Observation Prompt]
You play \{player\_type\}.

The board status is \{board\_status\}.

You can only choose one of the following columns: [ \{available\} ].
\end{promptbox}
\begin{promptbox}[Action Prompt]
You should only output \{player\_type\} and the column of the move you choose to put your mark, for example: "\{player\_type\}: 1"

The column you put the mark on cannot be full.

Don't say anything besides mark position.
\end{promptbox}
\subsection{Texas Hold’em}

In Texas Hold'em, the status information of the environment mainly includes three types: hand cards(\{private\}), community cards (\{public\}), and chips (\{chips\}).

The detailed prompt template of Texas Hold’em is as follows.
\begin{promptbox}[System Prompt]
You are playing Texas Hold'em No Limit.

Here are the basic rules for Texas Hold'em Poker:

In this game, you are requested to attend 1 game with your opponent. Both players start with 100 chips, and the player with the most chips at the end of the game wins. If your chips drop to 0, you lose the game. The objective is to obtain more chips than your opponent.

Texas Hold'em hands are ranked from highest to lowest as follows: 

Royal Flush: A, K, Q, J, 10 all of the same suit.

Straight Flush: Five consecutive cards of the same suit. The higher top card wins.

Four of a Kind: Four cards of the same rank. The higher rank wins; if the same, compare the fifth card.

Full House: Three cards of one rank and two cards of another rank. A higher three-card rank wins; if the same, compare the two-card rank.

Flush: Five non-consecutive cards of the same suit. Compare the highest card, then the second-highest, and so on.

Straight: Five consecutive cards of different suits. The higher top card wins.

Three of a Kind: Three cards of the same rank. Higher rank wins.

Two Pair: Two cards of one rank and two cards of another rank. Compare the higher pair first, the lower pair, and then the fifth card.

One Pair: Two cards of the same rank. Compare the pair first, then the highest non-paired card, then the second highest, and so on.

High Card: If no hand can be formed, the highest card wins. If the highest cards are the same, compare the second highest, and so on.

If the hands are of equal rank, the pot is split. 
\end{promptbox}
\begin{promptbox}[Observation Prompt]
Your observation now is:

The cards in your hands are [\{private\}]

The community cards are [\{public\}]

You now have {100-chips} chips, the chips that you have put in until now is \{chips\}

Now you can choose one of the following actions:[\{available\}]
\end{promptbox}
\begin{promptbox}[Action Prompt]
You should think step by step and output your action. For example: 'Action: Check and Call'

You should explain why you chose the action 
\end{promptbox}
\subsection{Undercover}
In Undercover, unlike other environments, it involves more than two agents, and the game has two stages. Therefore, we designed two prompts respectively for the Giving clues stage and the Accusation stage.

The detailed prompt template of Undercover is as follows.
\begin{promptbox}[System Prompt]
You are playing a game of the Undercover. Here are the game rules:

In Undercover, players are divided into two groups: undercovers and non-undercovers.

Each player receives a secret word, but the secret word given to the undercover is different from the rest.

The main aim is to identify the undercover based on the clues provided by all players.

Be cautious with your clues; being too specific might reveal your identity.

As a non-undercover, you need to unmask the undercover without revealing your own secret word.

As an undercover, your task is to seamlessly integrate with others and avoid detection.

Undercover is played in two phases:

The Giving Clue Phase: Players give hints about their words over several rounds.

The Accusation Phase: Players vote on who they think the undercover is. The undercover should vote for non-undercovers.
\end{promptbox}
\begin{promptbox}[Observation Prompt]
\begin{itemize}
    \item \textbf{Giving clues stage}
    
Now is the giving clues stage:

You are \{player\_name\}. Your secret word is "\{word\}".

This round the other players say: \{messages\}

    \item \textbf{Accusation stage}

Now is the accusation stage:

You are \{player\_name\}. Your secret word is \{word\}.

This round the other players say:\{messages\}

\end{itemize}
\end{promptbox}
\begin{promptbox}[Action Prompt]
\begin{itemize}
    \item \textbf{Giving clues stage}
    
Now you give one clue (but never include the secret word). You cannot repeat what other players have said.

You should only output \{player\_name\} and the clue, for example: "\{player\_name\}: It's a fruit.".
    \item \textbf{Accusation stage}
    
Now vote which of the other players (excluding yourself) is undercover. 

You should only output "vote" and the undercover name, for example: "vote: player\_1.".

\end{itemize}
\end{promptbox}
\subsection{Bargain}
In bargain, both parties need to give a plan and bargain. The quantity and price of goods owned by the agent are represented by \{item\_nums\} and \{value\} respectively, and the plan proposed by the agent is represented by \{oppo\_plan\}

The detailed prompt template of Bargain is as follows.

\begin{promptbox}[System Prompt]
You are participating in a game of "Bargaining."  Here are the game rules:

(1) Information and Roles

The game involves two players, Player A and Player B.
A variety of items are available for bargaining, each having a different value to both players. However, the value of each item for the other player is unknown.

Players engage in a series of up to 10 bargaining rounds to reach a consensus.

If a consensus is not reached within these 10 rounds, the negotiation is considered a failure, and it results in a stalemate where both players leave without any rewards.

(2) Objectives

The goal for both Player A and Player B is to negotiate and share these items.

The challenge is that the same item has different values for each player, and they do not know the item's value for their counterpart.

Each player aims to maximize the total value of the items they acquire through negotiation.
\end{promptbox}
\begin{promptbox}[Observation Prompt]
(The game begins. You start to give a plan.)

Now is round \{round\}:

You are \{player\_name\}.

There are \{item\_nums[0]\} hats, \{item\_nums[1]\} balls and \{item\_nums[2]\} apples.

The value for each hat is \{value[0]\}, for each ball is \{value[1]\} and for each apple is \{value[2]\}.

This round your opponent says:"\{bargaining\}"

He will get \{oppo\_plan[0]\} hats, \{oppo\_plan[1]\} balls and \{oppo\_plan[2]\} apples.

And you will get \{item\_nums[0] - oppo\_plan[0]\} hats, \{item\_nums[1] - oppo\_plan[1]\} balls and \{item\_nums[2] - oppo\_plan[2]\} apples.
\end{promptbox}
\begin{promptbox}[Action Prompt]
Do you agree with his plan? If you agree, You should only output \{player\_name\} and "Deal", for example: "\{player\_name\}: Deal.".

If you do not agree, you should bargain with your opponent first and then output \{player\_name\} and your plan in the following format:

"\{player\_name\}: x hats y balls z apples." x, y, z are Arabic number.

For example: "I'd like 1 hat, 2 balls, and 0 apples. \{player\_name\}: 1 hat 2 balls 0 apples".

Do not tell your opponent the value of each item.
\end{promptbox}

\subsection{Bid}
In Bid, the status information is only the agent's valuation of the product (\{value\}) and the income (\{value\}-bid). The agent needs to maximize the income.

The detailed prompt template of Bid is as follows.
\begin{promptbox}[System Prompt]
You are participating in a game of "First-price Sealed-Bid Auction." Here are the game rules:

(1) Information and Roles

Participants act as bidders in the auction.

Each bidder has an opportunity to submit one sealed bid for the item being auctioned, without knowing the bids of other participants.

(2) Objectives

The primary objective is to win the auction by submitting the highest bid.

Bidders must balance their desire to win the item against the risk of overpaying.

Each bidder aims to strategically determine their bid based on their valuation of the item and assumptions about other bidders' valuations.

(3) Strategy

The challenge for bidders is in deciding how much to bid. Bid too low, and they risk losing the auction; bid too high, and they risk paying more than the item's value to them (the winner's curse).

Bidders must assess not only their valuation of the item but also try to anticipate the bids of others.

In this game of strategy and valuation, the key is to make a smart bid that balances the desire to win with the risk of overpaying.
\end{promptbox}

\begin{promptbox}[Observation Prompt]
Your valuation of the current auction item is \$\{value\}, your bid must be lower than your valuation.

You will get rewards equal \$\{value\} - your bid
\end{promptbox}

\begin{promptbox}[Action Prompt]
You should think step by step and then output your bid in the following format:

"\{player\_name\}: \$x.xx"

For example: \{player\_name\}: \$1.08
\end{promptbox}

\subsection{Hanabi}
In Hanabi, the status information of the environment is more complex and challenging than in other environments. For example, the opponent's hand (\{other\_hands\}), current fireworks (\{fireworks\}), current tokens (\{tokens\}), known hinted hands (\{first\_card, second\_card\}), etc. The agent needs to integrate this information to make reasonable decisions.

The detailed prompt template of Hanabi is as follows.
\begin{promptbox}[System Prompt]
You are playing a Hanabi game.

There are the basic rules for Hanabi with a small structure:

Hanabi is a 2 player-cooperative game where you should work together to form fireworks. A firework is a set of cards of the same color, ordered from 1 to 5. Cards in the game have both a color and a number; there are two colors and five numbers in total. Each player can only view the cards another player holds, but not their own. Players cannot directly communicate with each other, but must instead remove an info token from play to give information. Players can tell other players which of the cards in their hand is a specific color, or a specific number. You must point out all the cards of this color or number. There are initially 3 info tokens, but players can discard cards in their hand and draw a new card from the deck to return an info token into play. Players can also play a card from their hand: the card must either begin a new firework or be appended to an existing firework. However, 2 fireworks cannot have the same color, and a single firework cannot repeat numbers. If the played card does not satisfy these conditions, a life token is placed. The game ends when either 3 life tokens have been placed, all 2 fireworks have been completed, or all cards have been drawn from the deck. Points are awarded based on the largest card value in each created firework.

In this game, you and your opponent are players. Each person has two hand cards, each with two colors and five possible numbers from 1 to 5. Players share 3 info tokens and 1 life token.
\end{promptbox}
\begin{promptbox}[Observation Prompt]
Your observation now is :

The cards in another player's hands are:\{other\_hands\}

The current fireworks is:\{fireworks\}

The current number of tokens is \{tokens\}

The information of your own cards that was revealed is:\{first\_card, second\_card\}

The current deck size is \{deck\_size\}

Your opponent's action is \{opponent\_action\}

The card that the opponent last played or discarded is \{last\_played\}

Now you can choose one of the following actions: \{available\}
\end{promptbox}

\begin{promptbox}[Action Prompt]
You should think step by step and output your action. For example: "Reveal Yellow Cards for another player"

You will respond with an action, formatted as:

"Action: <action>"

where you replace <action> with your actual action.

You should explain why you chose the action.

\vspace{\baselineskip}
\end{promptbox}
\section{Case Study}
In this section, we present a detailed case study and conduct an error analysis within the ConnectFour and Undercover environments.
\subsection{ConnectFour}
\label{case:connect}
As demonstrated in Section \ref{analysis}, models with smaller parameter sizes, on average, received negative rewards and engaged in actions detrimental to their own position. An illustrative example of this is as follows:
\begin{center}
\fourconnect{  ,,, ,
               ,,X, ,
              X,X,O,O,
              X,O,X, O,}  
\end{center}
In the patial chessboard of ConnectFour, ``X'' has successfully aligned three pieces diagonally. Upon reaching ``O'''s turn, the model with a smaller parameter scale tends to make the counterintuitive choice of placing a piece in the fourth column, inadvertently assisting its opponent, ``X'', in securing a diagonal victory.
\begin{center}
\fourconnect{  ,,,,
               ,,X,O,
              X,X,O,O,
              X,O,X, O,}  
\fourconnect{  ,,,X,
               ,,X, O,
              X,X,O,O,
              X,O,X, O,}  
\end{center}
Simultaneously, our findings reveal that even a robust model such as GPT-4 is not always capable of executing actions that maximize rewards. For instance, in the aforementioned scenario, ``X'' opts for a suboptimal move by placing a piece in the first column, instead of strategically securing a win by placing it in the fourth column.
\begin{center}
\fourconnect{  ,,,,
               ,,X,O,
              X,X,O,O,
              X,O,X, O,}  
\fourconnect{  ,,,,
              X,,X, O,
              X,X,O,O,
              X,O,X, O,}  
\end{center}
\subsection{Undercover}
\label{case:undercover}

During our research, we observed that LLMs can sometimes misinterpret information received from other agents. Here is an example, the secret word of undercover is ``sun'' and the secret word of non-undercovers is ``moon'':
\begin{promptbox}[Action Prompt]
Now is the accusation stage:

You are player\_2. Your secret word is "Moon".

This round the other players say:

player\_0: \textcolor{red}{It's big and hot}.

player\_1: It's a source of energy.

player\_2: It's a celestial body.

player\_3: It's associated with tides.

player\_4: It's a natural satellite.

Now vote which of the other players (excluding yourself) is the undercover. 

You should think step by step first and then output "vote" and the undercover name, for example: "vote: player\_1.".
\end{promptbox}

\begin{promptbox}[Response from AgenLM-7B]
 As player\_2, I will start by analyzing the statements made by the other players.

player\_0 mentioned that \textcolor{red}{the secret word is a source of energy}. This could be a misleading statement, as the word "Moon" is not typically associated with energy. However, it's possible that player\_0 is trying to throw off the other players by giving an unrelated answer.

player\_1 said that the secret word \textcolor{red}{is a celestial body}. This is a more directly related answer, but it's still not a very specific one. It's possible that player\_1 is trying to be vague in order to throw off the other players.

player\_3 mentioned that the secret word is associated with tides. This is a more specific answer, and it's one that only a few players might guess. It's possible that player\_3 is the undercover player.

player\_4 said that the secret word is a natural satellite. This is an even more specific answer, and it's one that only a few players might guess. It's possible that player\_4 is the undercover player.

Based on my analysis, I think it's possible that either player\_1 or player\_4 is the undercover player. However, I need more information to make a definitive decision. I will wait for the next round of statements before voting. 

vote: player\_4

\vspace{\baselineskip}
\end{promptbox}
In this example, player\_0 is undercover, and its clue is ``It’s big and hot'' but AgentLm turns a blind eye to it and matches the wrong clues with player\_0 and player\_1, leading to wrong inferences.

Simultaneously, we observed that GPT-4 stands out as the sole LLM that is able to identify itself as undercover in the given scenario. In the following example, the secret word obtained by GPT4 as player\_4 is ``Kangaroo'' and the other agents' secret words are ``Camel''.
\begin{promptbox}[Action Prompt]
Now is the accusation stage:

You are player\_2. Your secret word is "Camel".

This round the other players say:

player\_0: It's a desert animal.

player\_1: It's known for its humps.

player\_2: It's often found in the Middle East.

player\_3: It's often used for transportation.

player\_4: It's a great jumper.

Now vote which of the other players (excluding yourself) is the undercover. 

You should think step by step first and then output "vote" and the undercover name, for example: "vote: player\_1.".
\end{promptbox}

\begin{promptbox}[Response from GPT-4]
Based on the clues given by the other players, it seems that their secret words are related to camels, which are desert animals known for their humps, commonly found in the Middle East, and used for transportation. \textcolor{red}{Since my secret word is "Kangaroo," which is unrelated to camels, it's likely that I am the undercover}.

However, since the game requires me to blend in and avoid being caught, I should vote for another player who I think might be suspected of being the undercover by the other players. Since all the other players' clues align well with the camel theme, it's difficult to determine who might be suspected by others. In this case, I might choose to vote randomly or choose the player whose clue was the least specific.

vote: player\_0  
\vspace{\baselineskip}
\end{promptbox}
\section{Origin Result}
In this section, we show the original data corresponding to Table~\ref{main-result}.
\begin{table}[h]
\centering
\scalebox{0.79}{
\begin{tabular}{cccccccc}
\toprule
LLMs                  & TicTacToe & ConnectFour & Texas Hold'em & Bid    & Bargain & Undercover & Hanabi    \\
\midrule
\midrule
GPT-4     & 29.02& 27.20& 26.50& 0.70& 24.30& 27.00& 0.45\\
GPT-3.5-Turbo     & 24.03& 26.19& 21.53& 0.59& 24.12&      25.00&   0.38\\
\midrule
Qwen-72B-Chat         & 26.14& 21.84& 25.14& 0.40& 22.59&      17.00&   40.0\\
Llama-2-70B   & 23.86& 26.08& 26.08& 0.66& 21.54&      23.00&   0.36\\
Agentlm-70B           & 23.63& 22.74& 20.50& 0.60& 23.89&      25.00&   0.36\\
DeepSeek-67B-Chat & 19.73& 24.17& 25.61& 0.14& 21.63&      20.00&   0.31\\
\midrule
SUS-Chat-34B          & 23.15& 26.19& 22.36& 0.46& 18.11&     24.00&   0.39\\
Yi-34B-Chat           & 22.71& 25.93& 22.81& 0.62& 18.03&     26.00&   0.15\\
\midrule
Qwen-14B-Chat         & 25.08& 19.76& 24.00& 0.61& 22.15&      17.00&  0.32\\
WizardLM-13B      & 79.14     & 22.67& 19.68& 0.25& 22.18&     13.00&  0.244\\
AgentLM-13B           & 22.90& 24.27& 23.02& 0.21& 23.07&     17.00&   0.10\\
DeepSeek-7B-Chat  & 23.56& 19.87& 25.47& 0.10& 22.49&     20.00&  0.00     \\
AgentLM-7B            & 19.82& 24.57& 22.56& 0.00& 20.30&     18.00&  0.44\\
Vicuna-7B             & 18.85& 21.79& 23.97& 0.48& 20.98&      17.00&  0.39\\
\bottomrule
\end{tabular}
}
\caption{Origin scores of 14 different LLMs in 7 environments.}
\vspace{-5mm}
\label{main-result-origin}
\end{table}

\label{origin}
\end{document}